\documentclass{article}

\usepackage{arxiv}

\usepackage[utf8]{inputenc} 
\usepackage[T1]{fontenc}    
\usepackage{hyperref}       
\usepackage{url}            
\usepackage{booktabs}       
\usepackage{amsfonts}       
\usepackage{nicefrac}       
\usepackage{microtype}      
\usepackage{lipsum}

\usepackage{times}
\usepackage{helvet}
\usepackage{courier}
\usepackage[switch]{lineno}

\usepackage[colorinlistoftodos]{todonotes}
\usepackage{algorithm} 
\usepackage{algpseudocode}
\usepackage{amsmath}
\usepackage{amssymb}
\usepackage{multirow}
\usepackage{mathtools,mathptmx}
\usepackage{graphicx}
\usepackage{subcaption}
\usepackage{cleveref}
\usepackage{float}

\usepackage{xparse}
\crefformat{subfigure}{#2\onlyletter{#1}#3}
\crefrangeformat{subfigure}{(#3\onlyletter{#1}#4--#5\onlyletter{#2}#6)}
\crefmultiformat{subfigure}
  {(#2\onlyletter{#1}#3}
  {,~#2\onlyletter{#1}#3)}
  {,~#2\onlyletter{#1}#3)}
  {,~#2\onlyletter{#1}#3)}
\ExplSyntaxOn
\NewDocumentCommand{\onlyletter}{m}
 {
  \tl_set:Nx \l_tmpa_tl { #1 }
  \tl_item:Nn \l_tmpa_tl { -1 }
 }
\ExplSyntaxOff

\newcommand{\DTMM}{DTMM}
\newcommand{\MM}{Minimax}

\newcommand{\cmmnt}[1]{\ignorespaces}

\newcommand*{\rom}[1]{\uppercase\expandafter{\romannumeral #1\relax}}

\newcommand\footnoteref[1]{\protected@xdef\thefnmark{\ref{#1}}\footnotemark}

\title{A Generic Framework for Clustering Vehicle Motion Trajectories}

\author{
  Fazeleh S.Hoseini $^*$ \\
  Department of Computer Science and Engineering\\
  Chalmers University of Technology\\
  Göteborg, Sweden \\
  \texttt{fazeleh@chalmers.se} \\
   \And
 Sadegh Rahrovani $^*$\\
  Autonomous Drive Department in VolvoCars\\
   Göteborg, Sweden  \\
  \texttt{sadegh.rahrovani@volvocars.com} \\
     \And
  Morteza Haghir Chehreghani \\
  Department of Computer Engineering\\
  Chalmers University of Technology\\
  Göteborg, Sweden \\
  \texttt{morteza.chehreghani@chalmers.se} 
}

\begin{document}
\maketitle

\begin{abstract}
The development of autonomous vehicles requires having access to a large amount of data in the concerning driving scenarios. However, manual annotation of such driving scenarios is costly and subject to the errors in the rule-based trajectory labeling systems. To address this issue, we propose an effective non-parametric trajectory clustering framework consisting of five stages: (1)~aligning trajectories and quantifying their pairwise temporal dissimilarities, (2)~embedding the trajectory-based dissimilarities into a vector space, (3)~extracting transitive relations,  (4)~embedding the transitive relations into a new vector space,  and (5)~clustering the trajectories with an optimal number of clusters. We investigate and evaluate the proposed framework on a challenging real-world dataset consisting of annotated trajectories. We observe that the proposed framework achieves promising results, despite the complexity caused by having trajectories of varying length.  Furthermore, we extend the framework to validate the augmentation of the real dataset with synthetic data generated by a Generative Adversarial Network (GAN) where we examine whether the generated trajectories are consistent with the true underlying clusters.
\end{abstract}


\def\thefootnote{*}\footnotetext{These authors contributed equally to this work}
\section{Introduction}
Digitization and new business models have revolutionized many industries, where in the automotive sector, these forces are giving rise to four disruptive technology-driven mega trends that reinforce each other: diverse mobility, autonomous driving, electrification, and connectivity \cite{Mack-ensey&Company}. Autonomous vehicles are often considered to be one important direction for the future of transportation where the consumer acceptance and willingness to pay depends a lot on the development of safe and reliable technical solutions that can provide a premium user experience. However, in order to assess Autonomous Drive (AD) vehicle safety with confidence, statistical analyses have shown that fully autonomous vehicles would have to be driven for more than hundreds of millions of kilometers, and testing autonomous vehicles is seen as increasingly challenging \cite{rand}. This becomes of more challenge when we need to compare and evaluate different AD functionality design proposals/changes, since the same amount of distance needs to be driven again by the AD vehicle for the verification sign-off. Thus, driving billions of kilometers in the field with human safety-drivers are being questioned as an effective way to assure the reliability and safety of AD vehicles, and more advanced methods of safety and reliability assurance need to be adopted.

One alternative approach to address some of the above-mentioned issues is known as scenario-based verification, where a scenario database is created by collecting data and extracting driving scenarios/events that the ego vehicle is exposed to in the field. This database represents how traffic scenarios look like, where the naturalistic in-field driving behaviors  are quantified by scenario models. Once such a scenario database is developed, it can be used as a source for generating test cases for verification of the AD functionality in virtual and real driving test environments. 
Therefore, there is a push in the automotive industry for investigating and expanding the pallet of tools for establishing and maintaining the “traffic scenario database” up-to-date, for further verification purposes. Scenario trajectories are considered as sequences of the ego vehicle states interaction with the surrounding objects, and are obtained by processing the data collected by ego vehicle sensors. Relative movement of surrounding objects/vehicles, w.r.t. the ego vehicle, is considered as a \emph{scenario}.

Using collected raw sensor data, to label and extract driving scenarios of interest can be done via different approaches \cite{menzel2018scenarios}. Knowledge-based approaches enable us to extract scenarios based on explicit-rules and defining a scenario description with some threshold that needs to be set in advance. The main advantage with this approach is having insight and control on all steps, and being able to use our prior knowledge of driving scenarios. However, it might be subject to errors similar to many rule-based systems such as bias and missing unknown cases. Therefore, as a complement approach we investigate using exploratory data analysis and learning  tools (in particular clustering) to accelerate and verify the obtained scenario labels. The main benefit of this approach is to find some unknown scenarios, patterns or outlier trajectories in the data set, especially since explicit-rule based tools depend on the scenario threshold values and thus they might be prone to be missing or misclassifying those events. The main focus of this study is on unsupervised learning methods that  enable us to re-label the extracted scenario trajectories. This approach has, however, its own challenges. First of all, large amounts of data are being collected by leading car manufacturers to improve, verify, and validate self-driving systems. This needs to be analyzed efficiently in an unsupervised manner. Secondly, appropriate methods are required to handle clustering time-series of different length when analyzing driving trajectories. Finally, the clustering tool should be utilized to validate the quality of generated driving trajectories, compared to the original driving scenarios.

We propose a framework for unsupervised clustering of vehicle trajectories with varying lengths, called DTMM. We first extract the temporal relations, and then, we embed them into a low-dimensional vector space, wherein the underlying structures and clusters are well distinguishable.
Since such clusters have elongated and complex shapes and boundaries, thus at the next step, we compute the transitive relations in order to make the cluster boundaries well separated. Then, several clustering methods such as Gaussian Mixture Models (GMMs) might require feature or vector representations of the data. Therefore, we again embed the trajectories into a vector space, based on the pairwise transitive distances, such that the transitive relations are preserved. Finally, we apply the clustering method and analyze the results.

Furthermore, we extend the framework to validate the consistency of synthetic data generated by Generative Adversarial Networks (GANs) (e.g., RecAE-GAN and RecAE-WGAN), where we examine whether the generated trajectories are consistent with the real clusters.
In particular, in this study, we have a minority scenario cluster (the cut-in scenarios), and  we use the GAN models to augment this cluster.
Then, we employ DTMM to serve us as a validation tool to assess the quality of the generated trajectories. We investigate and evaluate the framework on a challenging real-world vehicle motion application and analyze the results.

\section{Data and Problem Description}

\begin{figure}[t]
\centering
\includegraphics[trim={0cm 0.4cm 0cm 1.5cm},clip,width=0.4\linewidth]{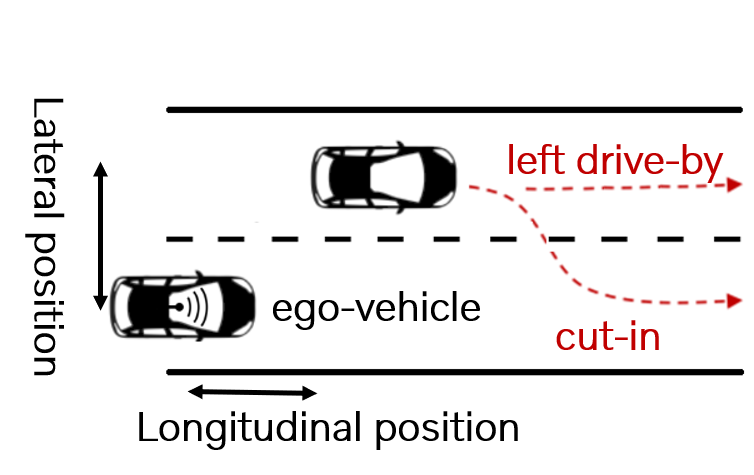}
\caption{ Examples of driving scenarios: cut-in and left drive-by.}\label{fig:cutin}
\end{figure}

Our driving scenario trajectory  data  consists  of information about surrounding vehicles, with respect to an ego vehicle over  different periods of time. Elements of a trajectory are described by two features: the relative lateral and longitudinal positions of the vehicle w.r.t. the ego vehicle, see Figure \ref{fig:cutin}. The trajectories can vary in length from 1 second up to 1 hour. The length depends on how long the object is tracked by the ego vehicle in the field of view (FoV). Here the main 
scenarios of interest are  cut-ins, left drive-by and right drive-by. Cut-in is defined as vehicles that approach the  ego  vehicle  from  the  left  lane and  overtake  the ego vehicle  by  switching  to  its  lane. The definition of a cut-in also  requires  the  vehicle  to  stay  in  front  of  the  ego  vehicle for at least 2 seconds. The length of the investigated trajectories varies between 3 to 7 seconds while the sampling frequency is 10 hertz. In this regard, the size of the considered sequences varies between 30 to 70. Scenarios are extracted from the logged data where this step is done with explicit rules defined by domain expert.

Different approaches have been developed for clustering of time-series, e.g., in  \cite{liao2005clustering,li2006coarse,macqueen1967some,salvador2007toward,Wang_2018,mhmm,berndt1994using,JohnMHMM}. Briefly, one issue with classical clustering methods such as $k$-means for time series clustering is that they require the data to belong to a metric space \cite{berndt1994using}. However, such a distance measure might not be well-defined for trajectories with different size. Padding the trajectories to make them have the same length would be a solution that is applied , e.g., in language processing. However, it is not an appropriate solution for driving scenario trajectories. Mapping the sequences to a feature space is a solution used by feature-based ML tools to handle sequentially of data \cite{liao2005clustering}, though most of them suffer from high computational complexity and very sensitive (hyper)parameters. 
Model-based clustering tools, such as Mixture of Hidden Markov Models (MHMM) \cite{JohnMHMM}, assign a generative model to each trajectory in order to alleviate the different length trajectories but the initialization of the generative model needs to be handled in a proper way and also require several nontrivial parameter settings. Using recurrent architectures deep learning models (Recurrent Auto-encoder), as feature extractor, is another solution while the trajectories could be clustered in the latent representation space of a fixed size \cite{Deepframework,WGAN}. Our framework will use some of the benefits of this paradigm for data augmentation.

\section{Trajectory Clustering Framework}\label{Sec:Sec3}
 
\begin{algorithm}[t]
	\caption{Trajectories clustering framework}
	\label{alg:overview}
    \hspace*{\algorithmicindent} \textbf{Input:}   Set of trajectories  $\mathbf{T}=\{\mathbf{t}_1, \mathbf{t}_2, ..., \mathbf{t}_K\}$\\
    \hspace*{\algorithmicindent} \textbf{Output:}  Set of cluster labels $\mathbf{L}=\{{l}_1, {l}_2, ..., {l}_K\}$
	\begin{algorithmic}[1]
	    
	    \State Trajectory alignment and quantification of pairwise deviations  $\mathbf{C} \in \mathbb{R}^{K\times K}$
	    \State Embedding trajectory-based dissimilarities as vectors $ \mathbf{V} \in \mathbb{R}^{K\times 2} $
	    \State Extraction of transitive relation $ \mathbf{M} \in \mathbb{R}^{K\times K} $
	    \State Embedding pairwise transitive relation to $ \mathbf{E} \in \mathbb{R}^{K\times d} $ where $d \leq K$
	    \State Clustering trajectories into an optimal number of clusters, labeled by  $\mathbf{L}=\{{l}_1, {l}_2, ..., {l}_K\}$ 
	\end{algorithmic} 
\end{algorithm}

\paragraph{\textit{\textbf{Framework overview}}} 
We are given a dataset $\mathbf{T} $ consisting of $K$ vehicle motion trajectories $\mathbf{T}=\{\mathbf{t}_1, \mathbf{t}_2, ..., \mathbf{t}_K\}$. Each trajectory $\mathbf{t}_i$, of length $n_i$,  represents a time series as a sequence of $n_i$ data points $\mathbf{t}_i=\{\mathbf{t}_i^{(1)}, \mathbf{t}_i^{(2)}, ..., \mathbf{t}_i^{(n_i)}\}$. Each $\mathbf{t}_i^{(j)}$ consists of two features: the lateral and the longitudinal positions of  the surrounding vehicle relative to the ego-vehicle.

We propose a generic framework for clustering the trajectories in $\mathbf{T}$. Algorithm \ref{alg:overview} describes the different steps of the framework. We first compute the pairwise dissimilarities between the variable-length time series and extract their temporal relations. Then, we embed the resultant pairwise dissimilarities into a vector space, while revealing the hidden clustering structures. Next, we extract the transitive relations between trajectories and again embed them into a vector space for better separability. Finally, in the vector space, we determine the optimal number of clusters and apply the clustering method. In the following, we motivate and explain each step of the proposed framework in detail.

\paragraph{\textit{\textbf{\rom{1}. Trajectory alignment and quantification of pairwise temporal dissimilarities}}}
The first step in our framework is to extract the temporal relations between the trajectories. For this purpose,  we employ the widely-used  Dynamic Time Warping (DTW) method \cite{DTWorigin}. DTW is frequently used to measure the dissimilarity between a pair of time series. It extracts the temporal relations disregarding up to a large extent the shift and distortion in time. Given two trajectories $\mathbf{t}_i$ and $\mathbf{t}_j$, the DTW algorithm first forms a  temporary distance matrix $\mathbf{D} \in \mathbb{R}^{n_i\times n_j}$,  that holds the pairwise dissimilarities between the elements  of  $\mathbf{t}_i$ and $\mathbf{t}_j$. It is recursively computed as
 \begin{equation}\label{eq:eq1}
\begin{aligned}
  \mathbf{D}_{n,m} =&  \ d(\mathbf{t}_i^{(n)}, \mathbf{t}_j^{(m)})
  & + \ min\big[  \mathbf{D}_{n-1,m},\ \mathbf{D}_{n,m-1},\ \mathbf{D}_{n-1,m-1}\big], 
\end{aligned}
\end{equation}
where $d(x,y) = \lvert x-y \rvert$, and  $ \mathbf{D}_{n,0} = \  \mathbf{D}_{0,m} = \ 0$. Since the elements of the trajectories  are not necessarily aligned, after obtaining $\mathbf{D}$, the algorithm discovers a path of aligned pairs of points, called an alignment path. This path is optimal in the sense that the sum over all pairs of point-wise misalignments is minimized.  We use this minimized cost, denoted by $\mathbf C_{ij}$, to define the dissimilarity between $\mathbf{t}_i$ and $\mathbf{t}_j$. Given $K$  trajectories, the alignment-cost matrix $\mathbf{C} \in \mathbb{R}^{K\times K}$ holds the pairwise misalignment costs between all the trajectories $\mathbf{t} \in \mathbf{T}$.

\paragraph{\textit{\textbf{\rom{2}. Embedding trajectory-based dissimilarities}}}
Our ultimate goal is to cluster the motion trajectories. Clusters are complex and possibly elongated structures or manifolds in data that usually lie in low-dimensional spaces. The reason is that in high dimensions the pairwise distances become almost equal and therefore the concept of cluster boundaries might not be well-defined anymore.
On the other hand, a vector representation of the pairwise dissimilarities $\mathbf C$ can give us a better understanding of the underlying clusters. Moreover, the pairwise dissimilarities $\mathbf C$ do not necessarily constitute a proper distance function, a condition that might be required by several clustering methods (e.g., those based on kernels or the mixture models). 
The reason is that the alignment cost might not fulfill the triangle inequality and thus does not qualify as a metric.
There are different methods for embedding pairwise dissimilarities of a set of elements into a vector space, such as Multidimensional Scaling (MDS) \cite{RePEc1938,MDS2348634} and t-Distributed Stochastic Neighbor Embedding (t-SNE) \cite{tsne}.
t-SNE aims to reveal underlying patterns and structures, with the common use of visualizing high-dimensional data in two or three dimensions.  
This method fits our purposes very well since it is capable of revealing global structure in the embedded vector space, while preserving the local mutual neighborhood relations.

In the first step, t-SNE computes a new set of pairwise (dis)similarities between all elements, that capture the degree to which they are mutual neighbors.
To start with, conditional probabilities $p_{j|i}$ are computed, measuring the probability that element $i$ would pick element $j$ as its neighbor.
The probabilities $p_{j|i}$ are set in proportion to the probability density under a Gaussian centered at element $i$, but are normalized such that they sum to $1$. 
Next, symmetric pairwise similarities are defined such that $p_{ij} \propto \frac{p_{j|i} + p_{i|j}}{2}$, and normalized such that $\sum_{i \neq j} p_{ij} = 1$, defining a joint distribution over all elements.
Finally, the elements are embedded as points in a low-dimensional vector space, such that their joint distribution $q_{ij}$, parameterized as a heavy-tailed student t-distribution with a single degree of freedom, will approximate $p_{ij}$ as good as possible.
The distance between a pair of embedded data points will now be short, if the original elements are similar neighbors, in the sense that the dissimilarity between these specific elements was relatively small.

Applying t-SNE on $\mathbf{C}$, we retrieve $K$ two-dimensional data points $\mathbf{V}_i$, one for each trajectory, which are then stacked in a matrix $ \mathbf{V} \in \mathbb{R}^{K\times 2}$.

\paragraph{\textit{\textbf{\rom{3}. Extraction of transitive relations}}}
As illustrated in Figure \ref{fig:result}, clusters usually have arbitrary elongated forms and variable shapes in low dimensions, such that for example they cannot be separated easily via hyperplanes. One way to improve the separability among different clusters is to take into account the transitive relations. If data point $a$ is close to data point $b$, $b$ is close to $c$, $c$ to ... to $z$, then we want to come up with a distance measure that yields a low distance between $a$ and $z$, even though their direct distance might be large. In this way, we aim at extracting the connectivity paths between the data points in order to extract elongated clusters with arbitrary shapes. On the other, since we assume an unsupervised learning setting, thus we prefer to perform this task without inducing any (critical) parameter, in order to obviate the need for a separate validation set that might not be available. 

Thus, we first represent the data 
by a graph $ G\left( \mathbf{O},\mathbf{W}\right)$ in which the nodes $\mathbf{O}$ correspond to the indices (of  the data points) and the edge weights are computed according to the pairwise squared Euclidean distances  between the respective feature vectors, i.e.,  $\mathbf{W}_{ij} = ||\mathbf{V}_i - \mathbf{V}_j||_2^2, \  \forall i,j \in \mathbf{O}$.

To compute such a transitive-aware distance measure, we can look for the smallest largest gap among all different paths between $i$ and $j$ on graph $\mathbf G$. For each particular path, we compute the largest gap (maximal edge weight), and then we choose the minimum gap of different paths. Therefore, this distance measure, known as Minimax or path-based distance \cite{FischerB03,chehreghani2019nonparametric} can be formulated as

\begin{equation}\label{eq:MM}
    \mathbf {M}_{ij} = \min_{r \in R_{ij}} \max_{e\in r} \mathbf W_{e}
\end{equation}
where $R_{ij}$ represents the set of all possible paths connecting $i$ and $j$ over graph  $G$. A path $r$ is characterized by the set of the consecutive edges $\{e\}$ on that, and $W_{e}$ presents the weight of edge $e$ (the weight between the two nodes at the two sides of $e$).

To compute the pairwise Minimax distances, we do not need to investigate all the possible paths between the nodes in $\mathbf G$. As discussed in \cite{ChehreghaniAAAI17}, pairwise Minimax distances over an arbitrary graph equal to their Minimax distances over any minimum spanning tree computed on that.
Thus, we first compute a minimum spanning tree on $\mathbf G$ using Prim's algorithm, and then employ the efficient dynamic programming method in \cite{ChehreghaniAAAI17} to compute the pairwise Minimax distances from the minimum spanning tree. Minimax distances have been successfully used in several learning tasks such as spectral clustering \cite{LittleMM20}, user profile completion \cite{ChehreghaniECIR17} and $k$-nearest neighbor search \cite{ChehreghaniSDM16,chehreghani2019nonparametric}.

\paragraph{\textit{\textbf{\rom{4}. Embedding pairwise transitive relation}}}
The previous steps yield a  matrix $\mathbf M$ of pairwise Minimax distances between all  trajectories.
Applying a clustering method typically requires a vector representation of the elements or converting the pairwise distances into a kernel matrix.
Computing a kernel might involve some free parameters that fixing proper values can be nontrivial \cite{Luxburg07}. On the other hand, some methods such as Gaussian Mixture Model (GMM) are applicable only to vectors (features). In general, vectors are the most basic way of data representation that many machine learning methods can be applied on.

Therefore, in this step we exploit the \emph{ultrametric} property of Minimax distances that ensures that the pairwise Minimax distances induce an $\mathcal{L}_2^2$ embedding \cite{chehreghani2019nonparametric}. In other words, there is a vector space wherein the squared Euclidean distances between the data points equal the pairwise Minimax distances in $\mathbf M$.
Then, after such a feasibility, we can use a method such as classical Multidimensional Scaling (MDS) to compute the embedding  $\mathbb{R}^d$, where $d \leq K$, and pairwise distances are perfectly preserved if $d=K$.
This method requires an eigenvalue decomposition, wherein one may use the elbow trick on the sorted eigenvalues to find the lowest possible $d$, such that $M$ is still approximated well. As we will discuss, finding a good $d$ is not critical at all.

\paragraph{\textit{\rom{5}. Clustering trajectories with an optimal number of clusters}}
Finally, we apply a clustering method to obtain the final clusters. Determining the optimal number of clusters in a dataset is a fundamental problem in clustering. In our work, we apply the silhouette method \cite{Silhouettes}, which optimizes for intra-cluster similarity and inter-cluster dissimilarity, by maximizing the so-called silhouette score. It varies between $-1$ and $+1$.
Thus, we perform the clustering multiple times with different numbers of clusters, and return the predicted labels $\mathbf{L}=\{{l}_1, {l}_2, ..., {l}_K\}$, corresponding to the solution that maximizes the silhouette score.

\section{Validation of GAN-Generated Trajectories}\label{sec:GAN_validation}

For scenario-based verification of autonomous vehicles, availability of a large and diverse scenario database is crucial.
However, some scenarios, in particular cut-in, happen less frequently compared to the other scenarios, making it harder to collect a diverse set of real in-field scenarios. Moreover, AD solutions need to be verified and validated not only based on the collected data from the field, but also based on similar perturbed trajectories that are not present in the data collection. To address such issues, we may employ Generative Adversarial Networks (GAN) \cite{GANbasic} to generate synthetic data by learning the statistical properties of the original data.

\begin{figure}[t]
\centering
\includegraphics[trim={5cm 0cm 53cm 0cm},clip,width=0.7\linewidth]{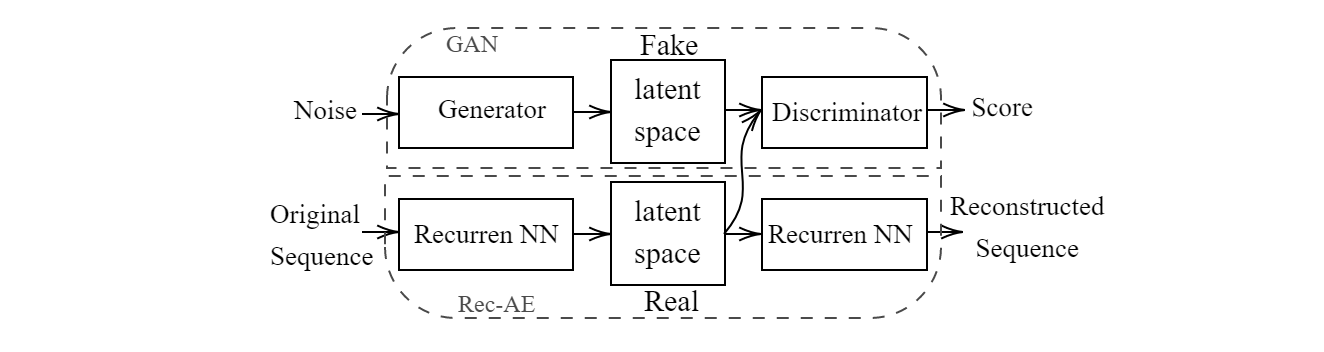}
\caption{Illustration of the RecAE-GAN model.}\label{fig:GAN}
\end{figure}

Due to the temporal/sequential nature of the trajectories, we use recurrent neural network models with GANs. 
In a Recurrent Auto-Encoder GAN (RecAE-GAN) \cite{WGAN}, illustrated in Figure \ref{fig:GAN}, a combination of recurrent neural networks and an auto-encoder is employed to encode the input set (sequence of elements) into a latent space. A generator network is trained to synthesize the latent space representation, while a discriminator network attempts to distinguish between the synthetic latent space representation generated by the generator network and the latent space representation of the original data.
In this work, we adapt the method developed in \cite{Deepframework}, wherein a feed-forward neural network is applied to estimate the length of a trajectory, in order to decode the trajectories from the latent space representation. We also employ recurrent AE with Wasserstein GAN (RecAE-WGAN) proposed in \cite{Deepframework} to generate scenario trajectories. It is worth mentioning that Wasserstein GAN introduced in \cite{arjovsky2017wasserstein}, is often used to improve the stability of learning by using the Wasserstein distance between the training data distribution and the generated data distribution, as the loss function.

Thereby, we adapt our clustering framework to investigate the  consistency of the synthetic trajectories generated by RecAE-GAN and RecAE-WGAN, and we 
examine whether the generated trajectories are consistent with the real clusters.
In this study, cut-in scenarios are rare compared to the other scenario trajectories. We, thus, in particular use the GAN models to generate scenarios of this type. 
Then, the proposed clustering framework can serve us as a validation tool to analyze the quality of the generated trajectories.

\section{Experimental Results} 
In this section, we 
describe the experimental setup, the different  baselines, the  evaluation criteria, and the clustering results on different datasets of real trajectories. We also investigate the generated  synthetic data and validate the consistency  with real data.

\subsection{Experimental setup}

We carry out our experiments on a set of annotated vehicle motion trajectories, where a surrounding vehicle is performing one out of three driving scenarios (cut-in, left drive-by, or right drive-by), and its motion trajectory is always relative to the ego vehicle. Our goal is to cluster the  trajectories such that each driving scenario corresponds to one single cluster.

To investigate our clustering framework, we extract six subsets from the full database, referred to as Set1-Set6, and presented in Table \ref{tab:sets}.
The number of trajectories per driving scenario, N, is either 256, 512, or 1024, depending on the experiment carried out.
Note that the main purpose for having Set4, Set5, and Set6 is to evaluate the performance when the real data is limited and thereby we study the impact of data augmentation through GANs. For this reason, Set5 and Set6 share the same real data as Set4, and only differ in the synthetic data augmented to Set4.

\begin{table}[t]
\caption{Clustering is evaluated on different evaluation sets with a varying range of trajectory lengths. The number of trajectories per driving scenario, N, is either 256, 512, or 1024.}
\centering
 \resizebox{0.7\linewidth}{!}{
\label{tab:sets}
\begin{tabular}{c||c|c|c||c|c||c}
\cline{2-6}
                            & \multicolumn{3}{c||}{Real}                                                                                                     & \multicolumn{2}{c||}{Synthetic}                                                                                                                        &                              \\ \cline{2-7} 
                            & \begin{tabular}[c]{@{}c@{}}Drive-by \\ Left\end{tabular} & \begin{tabular}[c]{@{}c@{}}Drive-by \\ Right\end{tabular} & Cut-in & \begin{tabular}[c]{@{}c@{}}Cut-in \\  (RecAE-GAN)\end{tabular} & \begin{tabular}[c]{@{}c@{}}Cut-in \\  (RecAE-WGAN)\end{tabular} &  \multicolumn{1}{c|}{\begin{tabular}[c]{@{}c@{}}Length\\ (sec)\end{tabular}} \\ \hline \hline
\multicolumn{1}{|c||}{Set 1} & N                                                        & N                                                         & N      & -                                                                      & -                                                                      & \multicolumn{1}{c|}{3-4} \\ \hline
\multicolumn{1}{|c||}{Set 2} & N                                                        & N                                                         & N      & -                                                                      & -                                                                      & \multicolumn{1}{c|}{4-6} \\ \hline
\multicolumn{1}{|c||}{Set 3} & N                                                        & N                                                         & N      & -                                                                      & -                                                                      & \multicolumn{1}{c|}{3-7} \\ \hline
\multicolumn{1}{|c||}{Set 4}  & N                                                        & N                                                         & N/2    & -                                                                    & -                                                                    & \multicolumn{1}{c|}{3-7} \\ \hline
\multicolumn{1}{|c||}{Set 5} & N                                                        & N                                                         & N/2    & N/2                                                                    & -                                                                      & \multicolumn{1}{c|}{3-7} \\ \hline
\multicolumn{1}{|c||}{Set 6} & N                                                        & N                                                         & N/2    & -                                                                      & N/2                                                                    & \multicolumn{1}{c|}{3-7} \\ \hline
\end{tabular}
}
\end{table}

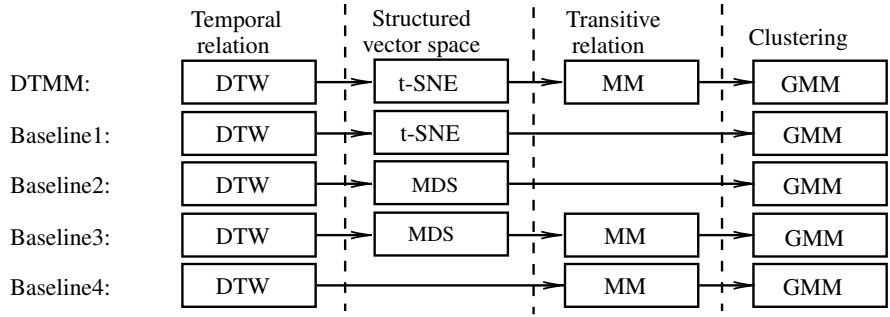
\begin{figure}[b]
  \begin{center}
\scalebox{.9}{

\tikzset{every picture/.style={line width=1pt}} 

\begin{tikzpicture}[x=0.8pt,y=0.45pt,yscale=-1,xscale=1]

\draw   (134,141) -- (204,141) -- (204,181) -- (134,181) -- cycle ;
\draw   (434,141) -- (504,141) -- (504,181) -- (434,181) -- cycle ;
\draw   (235,140) -- (305,140) -- (305,180) -- (235,180) -- cycle ;
\draw    (204.5,161) -- (231.5,161) ;
\draw [shift={(233.5,161)}, rotate = 180] [color={rgb, 255:red, 0; green, 0; blue, 0 }  ][line width=0.75]    (10.93,-3.29) .. controls (6.95,-1.4) and (3.31,-0.3) .. (0,0) .. controls (3.31,0.3) and (6.95,1.4) .. (10.93,3.29)   ;
\draw    (304.5,161) -- (432.5,161) ;
\draw [shift={(434.5,161)}, rotate = 180] [color={rgb, 255:red, 0; green, 0; blue, 0 }  ][line width=0.75]    (10.93,-3.29) .. controls (6.95,-1.4) and (3.31,-0.3) .. (0,0) .. controls (3.31,0.3) and (6.95,1.4) .. (10.93,3.29)   ;
\draw   (134,188) -- (204,188) -- (204,228) -- (134,228) -- cycle ;
\draw   (434,188) -- (504,188) -- (504,228) -- (434,228) -- cycle ;
\draw   (235,187) -- (305,187) -- (305,227) -- (235,227) -- cycle ;
\draw    (204.5,208) -- (231.5,208) ;
\draw [shift={(233.5,208)}, rotate = 180] [color={rgb, 255:red, 0; green, 0; blue, 0 }  ][line width=0.75]    (10.93,-3.29) .. controls (6.95,-1.4) and (3.31,-0.3) .. (0,0) .. controls (3.31,0.3) and (6.95,1.4) .. (10.93,3.29)   ;
\draw    (304.5,208) -- (432.5,208) ;
\draw [shift={(434.5,208)}, rotate = 180] [color={rgb, 255:red, 0; green, 0; blue, 0 }  ][line width=0.75]    (10.93,-3.29) .. controls (6.95,-1.4) and (3.31,-0.3) .. (0,0) .. controls (3.31,0.3) and (6.95,1.4) .. (10.93,3.29)   ;
\draw   (134,236) -- (204,236) -- (204,276) -- (134,276) -- cycle ;
\draw   (335,236) -- (405,236) -- (405,276) -- (335,276) -- cycle ;
\draw   (434,236) -- (504,236) -- (504,276) -- (434,276) -- cycle ;
\draw   (235,235) -- (305,235) -- (305,275) -- (235,275) -- cycle ;
\draw    (204.5,256) -- (231.5,256) ;
\draw [shift={(233.5,256)}, rotate = 180] [color={rgb, 255:red, 0; green, 0; blue, 0 }  ][line width=0.75]    (10.93,-3.29) .. controls (6.95,-1.4) and (3.31,-0.3) .. (0,0) .. controls (3.31,0.3) and (6.95,1.4) .. (10.93,3.29)   ;
\draw    (304.5,256) -- (331.5,256) ;
\draw [shift={(333.5,256)}, rotate = 180] [color={rgb, 255:red, 0; green, 0; blue, 0 }  ][line width=0.75]    (10.93,-3.29) .. controls (6.95,-1.4) and (3.31,-0.3) .. (0,0) .. controls (3.31,0.3) and (6.95,1.4) .. (10.93,3.29)   ;
\draw    (405.5,256) -- (432.5,256) ;
\draw [shift={(434.5,256)}, rotate = 180] [color={rgb, 255:red, 0; green, 0; blue, 0 }  ][line width=0.75]    (10.93,-3.29) .. controls (6.95,-1.4) and (3.31,-0.3) .. (0,0) .. controls (3.31,0.3) and (6.95,1.4) .. (10.93,3.29)   ;
\draw   (134,283) -- (204,283) -- (204,323) -- (134,323) -- cycle ;
\draw   (335,283) -- (405,283) -- (405,323) -- (335,323) -- cycle ;
\draw   (434,283) -- (504,283) -- (504,323) -- (434,323) -- cycle ;
\draw    (204.5,303) -- (332.5,303) ;
\draw [shift={(334.5,303)}, rotate = 180] [color={rgb, 255:red, 0; green, 0; blue, 0 }  ][line width=0.75]    (10.93,-3.29) .. controls (6.95,-1.4) and (3.31,-0.3) .. (0,0) .. controls (3.31,0.3) and (6.95,1.4) .. (10.93,3.29)   ;
\draw    (405.5,303) -- (432.5,303) ;
\draw [shift={(434.5,303)}, rotate = 180] [color={rgb, 255:red, 0; green, 0; blue, 0 }  ][line width=0.75]    (10.93,-3.29) .. controls (6.95,-1.4) and (3.31,-0.3) .. (0,0) .. controls (3.31,0.3) and (6.95,1.4) .. (10.93,3.29)   ;
\draw   (134,93) -- (204,93) -- (204,133) -- (134,133) -- cycle ;
\draw   (335,93) -- (405,93) -- (405,133) -- (335,133) -- cycle ;
\draw   (434,93) -- (504,93) -- (504,133) -- (434,133) -- cycle ;
\draw   (235,92) -- (305,92) -- (305,132) -- (235,132) -- cycle ;
\draw    (204.5,113) -- (231.5,113) ;
\draw [shift={(233.5,113)}, rotate = 180] [color={rgb, 255:red, 0; green, 0; blue, 0 }  ][line width=0.75]    (10.93,-3.29) .. controls (6.95,-1.4) and (3.31,-0.3) .. (0,0) .. controls (3.31,0.3) and (6.95,1.4) .. (10.93,3.29)   ;
\draw    (304.5,113) -- (331.5,113) ;
\draw [shift={(333.5,113)}, rotate = 180] [color={rgb, 255:red, 0; green, 0; blue, 0 }  ][line width=0.75]    (10.93,-3.29) .. controls (6.95,-1.4) and (3.31,-0.3) .. (0,0) .. controls (3.31,0.3) and (6.95,1.4) .. (10.93,3.29)   ;
\draw    (405.5,113) -- (432.5,113) ;
\draw [shift={(434.5,113)}, rotate = 180] [color={rgb, 255:red, 0; green, 0; blue, 0 }  ][line width=0.75]    (10.93,-3.29) .. controls (6.95,-1.4) and (3.31,-0.3) .. (0,0) .. controls (3.31,0.3) and (6.95,1.4) .. (10.93,3.29)   ;
\draw  [dash pattern={on 4.5pt off 4.5pt}]  (318.5,43) -- (318.5,330) ;
\draw  [dash pattern={on 4.5pt off 4.5pt}]  (219.5,40) -- (219.5,327) ;
\draw  [dash pattern={on 4.5pt off 4.5pt}]  (417.5,40) -- (417.5,327) ;

\draw (150,152) node [anchor=north west][inner sep=0.75pt]   [align=left] {DTW};
\draw (247,150) node [anchor=north west][inner sep=0.75pt]   [align=left] {t-SNE};
\draw (448,152) node [anchor=north west][inner sep=0.75pt]   [align=left] {GMM};
\draw (42,152) node [anchor=north west][inner sep=0.75pt]   [align=left] {Baseline1:};
\draw (150,199) node [anchor=north west][inner sep=0.75pt]   [align=left] {DTW};
\draw (253,199) node [anchor=north west][inner sep=0.75pt]  [font=\small] [align=left] {MDS};
\draw (448,200) node [anchor=north west][inner sep=0.75pt]   [align=left] {GMM};
\draw (42,199) node [anchor=north west][inner sep=0.75pt]   [align=left] {Baseline2:};
\draw (150,247) node [anchor=north west][inner sep=0.75pt]   [align=left] {DTW};
\draw (251,245) node [anchor=north west][inner sep=0.75pt]  [font=\small] [align=left] {MDS};
\draw (449,248) node [anchor=north west][inner sep=0.75pt]   [align=left] {GMM};
\draw (354,247) node [anchor=north west][inner sep=0.75pt]   [align=left] {MM};
\draw (42,247) node [anchor=north west][inner sep=0.75pt]   [align=left] {Baseline3:};
\draw (150,294) node [anchor=north west][inner sep=0.75pt]   [align=left] {DTW};
\draw (448,294) node [anchor=north west][inner sep=0.75pt]   [align=left] {GMM};
\draw (354,294) node [anchor=north west][inner sep=0.75pt]   [align=left] {MM};
\draw (42,294) node [anchor=north west][inner sep=0.75pt]   [align=left] {Baseline4:};
\draw (150,104) node [anchor=north west][inner sep=0.75pt]   [align=left] {DTW};
\draw (246,103) node [anchor=north west][inner sep=0.75pt]   [align=left] {t-SNE};
\draw (447,104) node [anchor=north west][inner sep=0.75pt]   [align=left] {GMM};
\draw (353,104) node [anchor=north west][inner sep=0.75pt]   [align=left] {MM};
\draw (42,104) node [anchor=north west][inner sep=0.75pt]   [align=left] {DTMM:};
\draw (137,44) node [anchor=north west][inner sep=0.75pt]   [align=left] {Temporal\\ \ relation};
\draw (218.5,43) node [anchor=north west][inner sep=0.75pt]   [align=left] {\begin{minipage}[lt]{63.383412pt}\setlength\topsep{0pt}
\begin{center}
Structured\\ vector space
\end{center}

\end{minipage}};
\draw (334,44) node [anchor=north west][inner sep=0.75pt]   [align=left] {Transitive\\ \ relation};
\draw (430,60) node [anchor=north west][inner sep=0.75pt]   [align=left] {Clustering};

\end{tikzpicture}

}
  \end{center}
  \caption{Overview of \DTMM\ and baseline methods.}
  \label{fig:baselines}
\end{figure}

We compare our clustering framework \DTMM, with four baselines, where an overview of all methods is presented in Figure \ref{fig:baselines}.
This comparison enables us to draw conclusions about the importance of the different steps in \DTMM, each of which is described in Section \ref{Sec:Sec3}. In Baseline1 and Baseline2, we do not make use of \MM, effectively disregarding the extraction of transitive relations between trajectories, as explained in Step (\rom{3}).
Whenever \MM\ is used, i.e. for DTMM as well as for Baseline3 and Baseline4, we also perform an embedding step before clustering, as explained in Step (\rom{4}). In Baseline2 and Baseline3, we modify the embedding of trajectory-based distances explained in Step (\rom{2}), replacing t-SNE with MDS. Instead of classical MDS, we use non-metric Multidimensional Scaling (nMDS), since the DTW dissimilarity does not necessarily fulfill a metric.  More specifically we use the  SMACOF algorithm \cite{Leeuw77applicationsof}. In Baseline4 we disregard this step, which is possible since \MM\ is applied directly on a distance matrix, and an embedding is not technically needed at this step.

\subsection{Clustering results and analysis}

We perform the methods introduced in Figure \ref{fig:baselines} on the datasets described in Table \ref{tab:sets} to examine our framework. We present results where a GMM is used for clustering, while pointing out that additional experiments with $k$-means yield consistent results. We have access to ground truth for our datasets, so we evaluate the result by three commonly used clustering criteria: rand index score \cite{rand} (RI), mutual information \cite{mutualinfo} (MI), and v-measure \cite{vmeasure} (VM) that respectively demonstrate the similarity, agreement, and homogeneity of estimated labels and ground truth.

\begin{table}[H]
\caption{
    Clustering results of \DTMM\ and baseline methods.
}\label{tab:result}
\centering 
\resizebox{0.5\linewidth}{!}{
\begin{tabular}{c|c|c|ccccc}
 & \multicolumn{1}{|c|}{Set} & {Metric} & B1  & B2 & B3 & B4 & \DTMM \\

     \hline  \hline
  
\multirow{25}{*}{\rotatebox{90}{512 points}}
 & \multirow{4}{*}{Set1} 
  & RI & 0.950   & 0.737  & 0.641 & 0.902 &   \textbf{1.0} \\
 & & MI & 0.938 &  0.737 & 0.711 & 0.896 &  \textbf{1.0} \\
 & & VM & 0.938  & 0.738 &  0.740 & 0.899 &  \textbf{1.0} \\
 & & SS &  0.514 & 0.684 &  0.961   &  0.867&  \textbf{0.996} \\
  \cline{2-8}
  & \multirow{4}{*}{Set2}  
  & RI & \textbf{1.0}  & 0.996 & 0.956 &  0.889 & \textbf{1.0} \\
 & & MI & \textbf{1.0}  & 0.992 & 0.943 &  0.883 & \textbf{1.0} \\
 & & VM & \textbf{1.0}  & 0.992 & 0.944 &  0.885 & \textbf{1.0} \\
 & & SS & 0.630         & 0.485 & 0.871 & 0.857  & \textbf{0.940} \\
 \cline{2-8}
& \multirow{4}{*}{Set3} 
   & RI & \textbf{1.0}  & 0.994 & 0.921 &  0.943 & \textbf{1.0} \\
 & & MI & \textbf{1.0}  & 0.897 & 0.906 &  0.931 & \textbf{1.0} \\
 & & VM & \textbf{1.0}  & 0.898 & 0.907 &  0.932 & \textbf{1.0} \\
 & & SS & 0.650 &  0.746 & 0.889 & 0.889  & \textbf{0.976} \\
  \cline{2-8}
 & \multirow{4}{*}{Set4} 
   & RI & 0.987 & 0.991 &  0.914 & 0.981  & \textbf{1.0} \\
 & & MI & 0.975 & 0.986 &  0.887 & 0.966  & \textbf{1.0} \\
 & & VM & 0.977 & 0.987 &  0.895 & 0.968  & \textbf{1.0} \\
 & & SS & 0.487 & 0.492 &  0.898 & 0.830  & \textbf{0.980} \\
  \cline{2-8}
 & \multirow{4}{*}{Set5} 
  & RI &  0.546 & 0.986 & 0.899 &  0.901  & \textbf{1.0} \\
 & & MI & 0.602 & 0.977 & 0.882 &  0.895  & \textbf{1.0} \\
 & & VM & 0.611 & 0.978 & 0.884 &  0.897  & \textbf{1.0} \\
 & & SS & 0.456 & 0.460 & 0.849 &  0.832  & \textbf{0.925} \\
 \cline{2-8}
  & \multirow{4}{*}{Set6} 
   & RI & 0.735  & 0.989  &  0.893   &  0.982  & \textbf{1.0} \\
 & & MI & 0.737  & 0.978  &  0.885   &  0.973  & \textbf{1.0}\\
 & & VM & 0.744  & 0.978  &  0.887   &  0.973  & \textbf{1.0} \\ 
  & & SS & 0.428         &  0.471 & 0.848  &  0.802  & \textbf{0.977} \\
  \hline \hline

\multirow{25}{*}{\rotatebox{90}{1024 points}}& \multirow{3}{*}{Set1}  
   & RI &  \textbf{1.0}   & 0.965 & 0.789  &0.896 & \textbf{1.0}  \\
 & & MI &  \textbf{1.0}  & 0.955 & 0.768  & 0.858 & \textbf{1.0} \\
 & & VM &  \textbf{1.0}  & 0.953&0.766 & 0.857  & \textbf{1.0}  \\
  & & SS & 0.508 &0.698 & 0.755&  0.658 & \textbf{0.936} \\

 \cline{2-8}
& \multirow{4}{*}{Set2} 

   & RI & \textbf{1.0}  & 0.960 & 0.811 &  0.936 & \textbf{1.0} \\
 & & MI & \textbf{1.0}  & 0.960 & 0.807 &  0.925 & \textbf{1.0} \\
 & & VM & \textbf{1.0}  & 0.958 &0.806  &  0.925 & \textbf{1.0} \\
 & & SS & 0.515        & 0.564 & 0.876 & 0.769  & \textbf{0.997} \\
  \cline{2-8}
 & \multirow{4}{*}{Set3} 
   & RI & 0.989 & 0.922 &  0.918 &  0.896 & \textbf{1.0}  \\
 & & MI & 0.981 & 0.902 &  0.908 &  0.891 & \textbf{1.0} \\
 & & VM & 0.981 & 0.903 &  0.909 &  0.893 & \textbf{1.0}  \\
 & & SS & 0.620 & 0.754 &  0.888 &  0.755 & \textbf{0.998} \\
  \cline{2-8}
 & \multirow{4}{*}{Set4} 
   & RI & 0.991  & 0.980 &  0.956 & 0.942  & \textbf{1.0}  \\
 & & MI & 0.982  & 0.964 &  0.930 & 0.918  & \textbf{1.0} \\
 & & VM & 0.983  & 0.967 &  0.934 & 0.924  & \textbf{1.0}  \\
 & & SS & 0.602  & 0.580 &  0.894 & 0.778  & \textbf{0.993} \\
 
 \cline{2-8}
  & \multirow{4}{*}{Set5} 
   & RI & 0.991  & 0.914 &  0.945 & 0.890 & \textbf{1.0} \\
 & & MI & 0.982  & 0.901 &  0.932 & 0.885 & \textbf{1.0} \\
 & & VM & 0.982  & 0.902 &  0.932 & 0.887 & \textbf{1.0}  \\
 & & SS & 0.558  & 0.748& 0.860&  0.687 & \textbf{0.998} \\
 \cline{2-8}
  & \multirow{4}{*}{Set6} 
   & RI & \textbf{1.0}   &  0.939  &   0.944 &  0.972 &  \textbf{1.0}  \\
 & & MI & \textbf{1.0}   &  0.920  &   0.930 &  0.960  & \textbf{1.0} \\
 & & VM & \textbf{1.0}   &  0.920  &   0.931 &  0.960  & \textbf{1.0}  \\
 & & SS & 0.527 & 0.779 & 0.884 & 0.796  & \textbf{0.998} \\
 
\end{tabular}
}
\end{table}

\begin{figure}[H] \centering
\begin{minipage}[b]{0.60\linewidth}
\begin{subfigure}[b]{\textwidth}
    \centering
    \captionsetup{justification=centering}
        \includegraphics[trim={0cm 10cm 0cm 0cm},clip,width=0.7\textwidth]{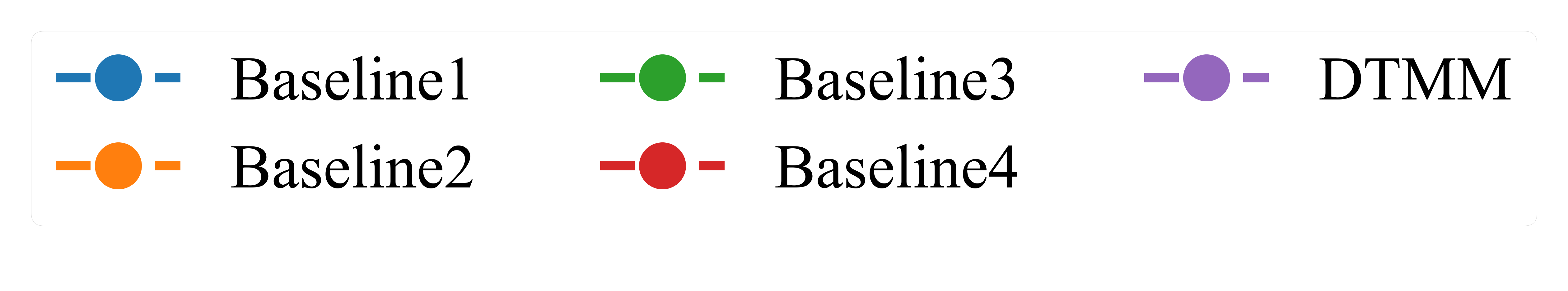}
    \end{subfigure}
    \begin{subfigure}[b]{0.49\textwidth}
    \centering
    \captionsetup{justification=centering}
        \includegraphics[trim={0cm 0cm 1cm 1.5cm},clip,width=\textwidth]{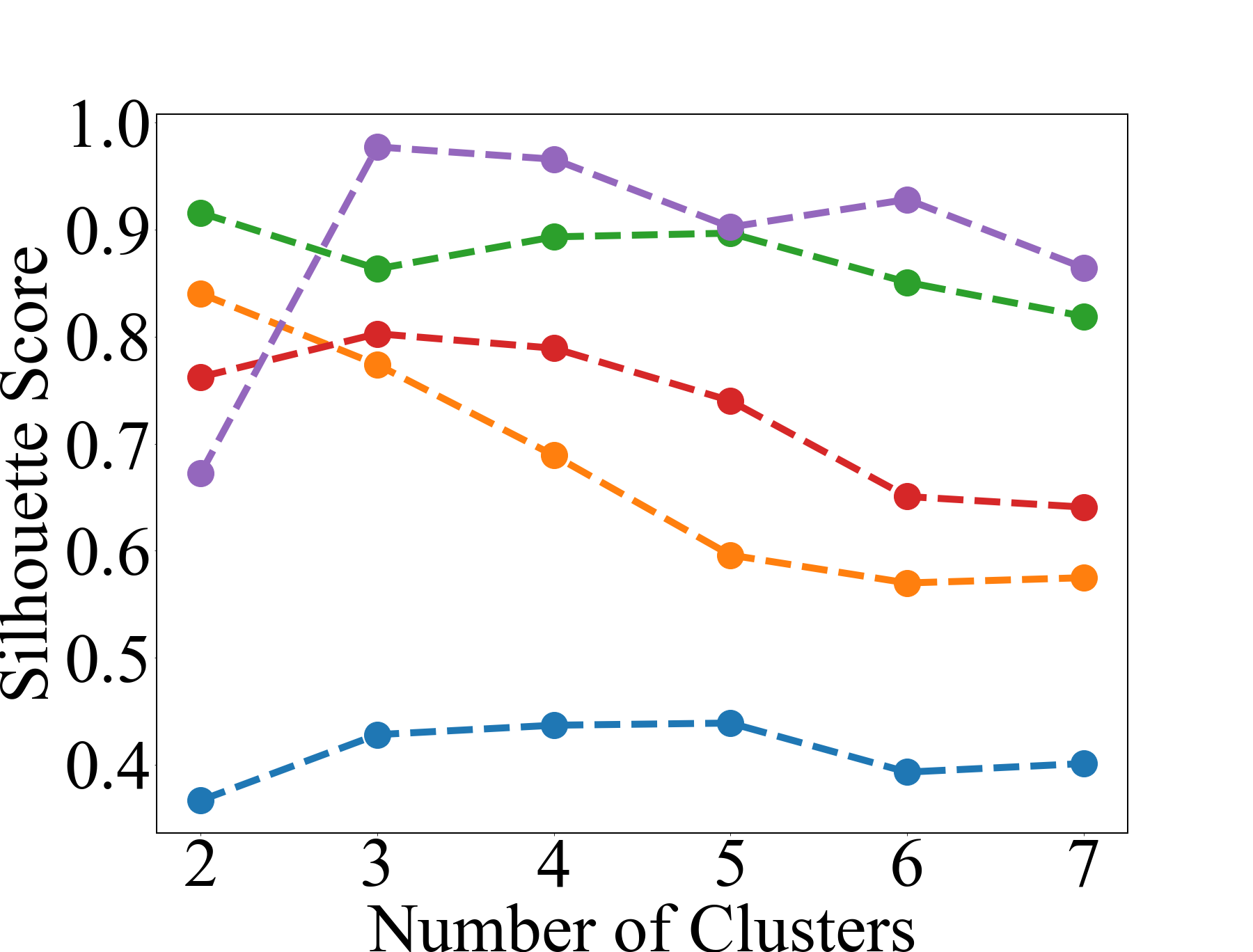}
         \caption{ \footnotesize{Silhouette Score} }
         \label{fig:Sallihoutte}
    \end{subfigure}
    \begin{subfigure}[b]{0.49\textwidth}
    \centering
    \captionsetup{justification=centering}
       \includegraphics[trim={0cm 0cm 1cm 1.5cm},clip,width=\textwidth]{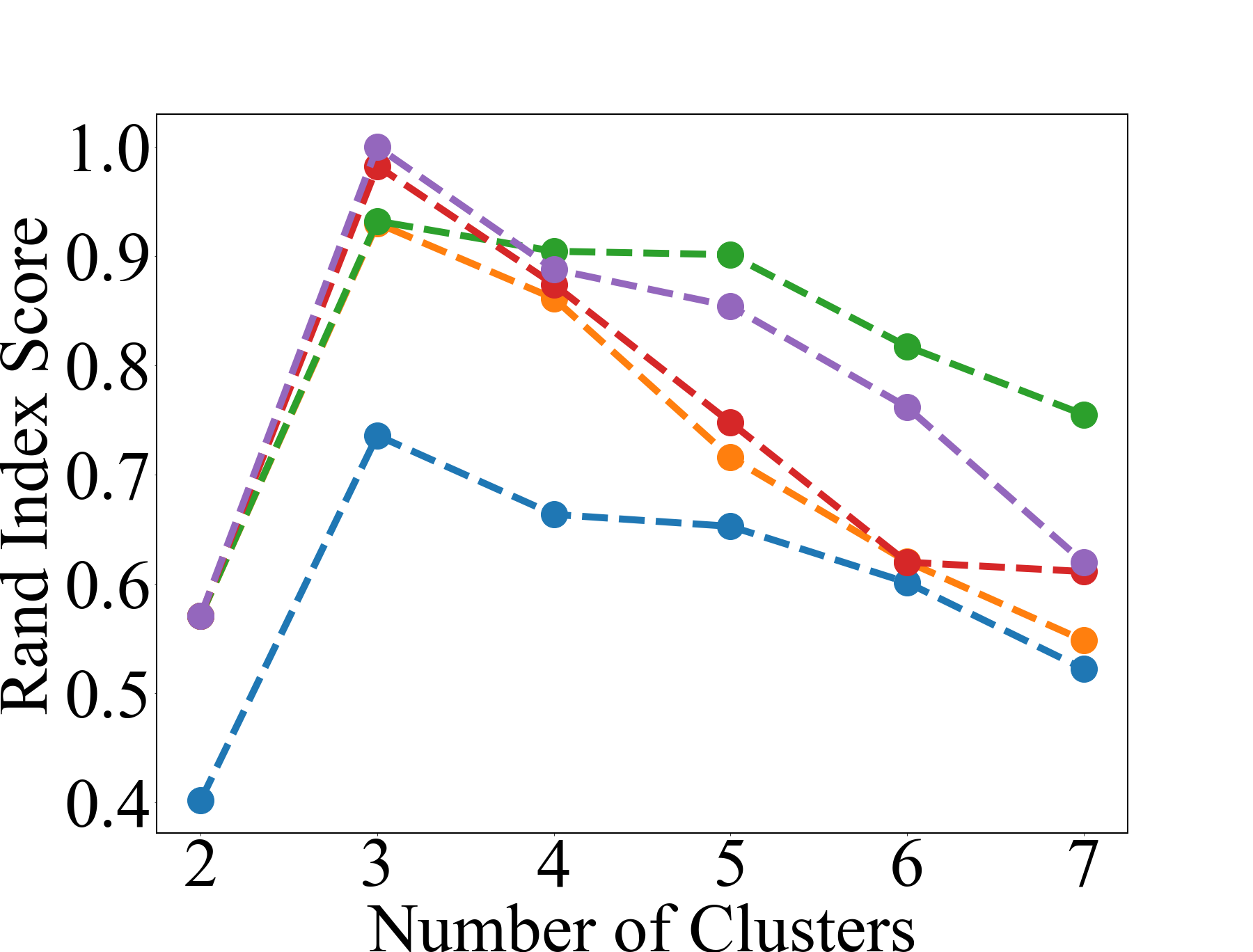}
         \caption{\footnotesize {Rand Index Score}}
         \label{fig:randindex}
    \end{subfigure}
\begin{subfigure}[b]{0.49\textwidth}
    \centering
    \captionsetup{justification=centering}
        \includegraphics[width=\textwidth]{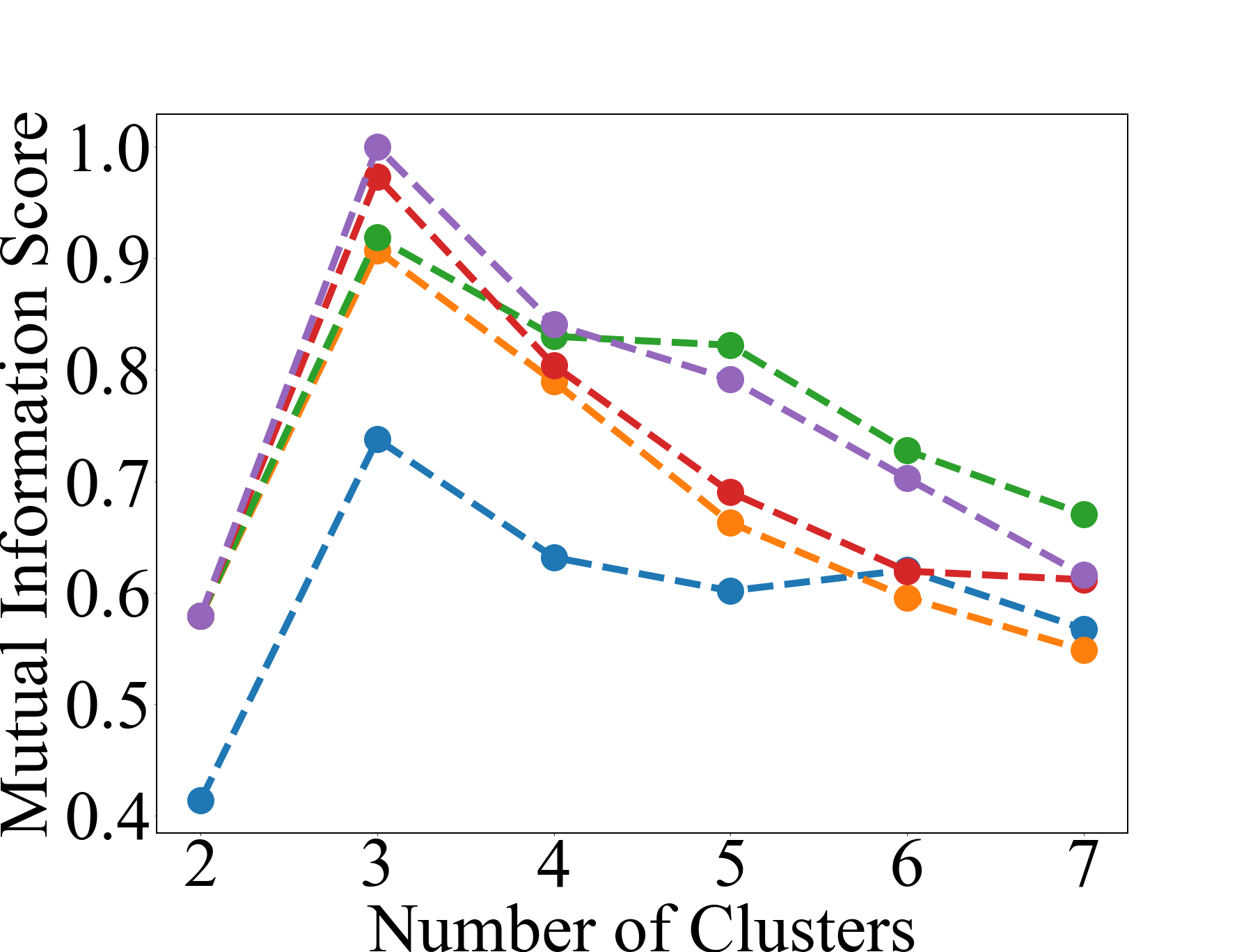}
         \caption{ \footnotesize{Mutual Info. Score} }
         \label{fig:mutualinfo}
    \end{subfigure}
    \begin{subfigure}[b]{0.49\textwidth}
    \centering
    \captionsetup{justification=centering}
       \includegraphics[width=\textwidth]{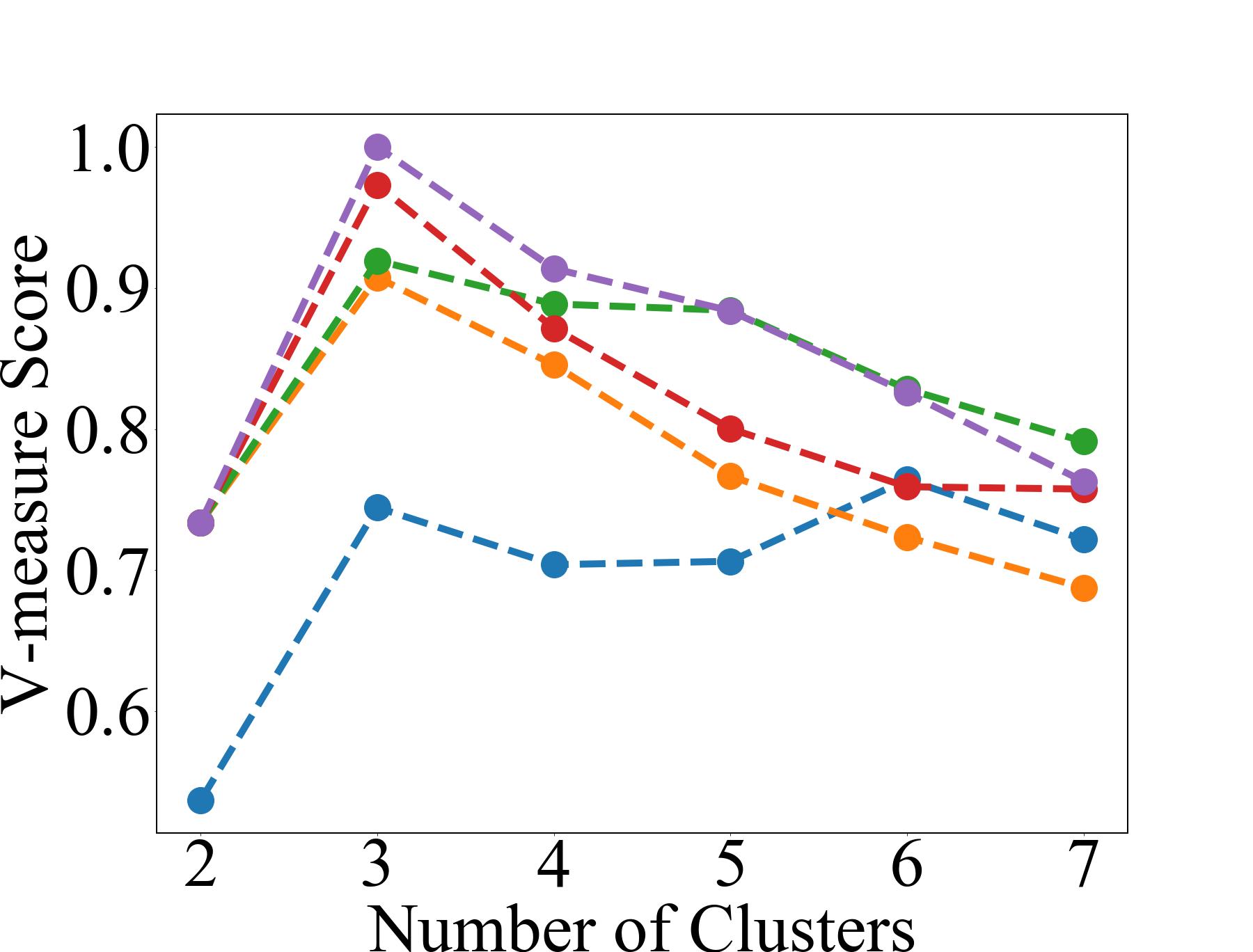}
         \caption{\footnotesize{V-measure Score} }
         \label{fig:vmeasure}
    \end{subfigure}
        \captionsetup{singlelinecheck = false, justification=justified}
    \end{minipage}\caption{Silhouette, rand index, mutual information and v-measure scores for Set6, $N=512$.}\label{fig:optimalclustering}
\end{figure}

\begin{figure}[b]
\begin{minipage}[b]{0.9\linewidth} \centering
    \begin{subfigure}[b]{0.3\textwidth}
    \centering
    \captionsetup{justification=centering}
        \includegraphics[trim={0cm 3cm 3.5cm 10cm},width=\textwidth]{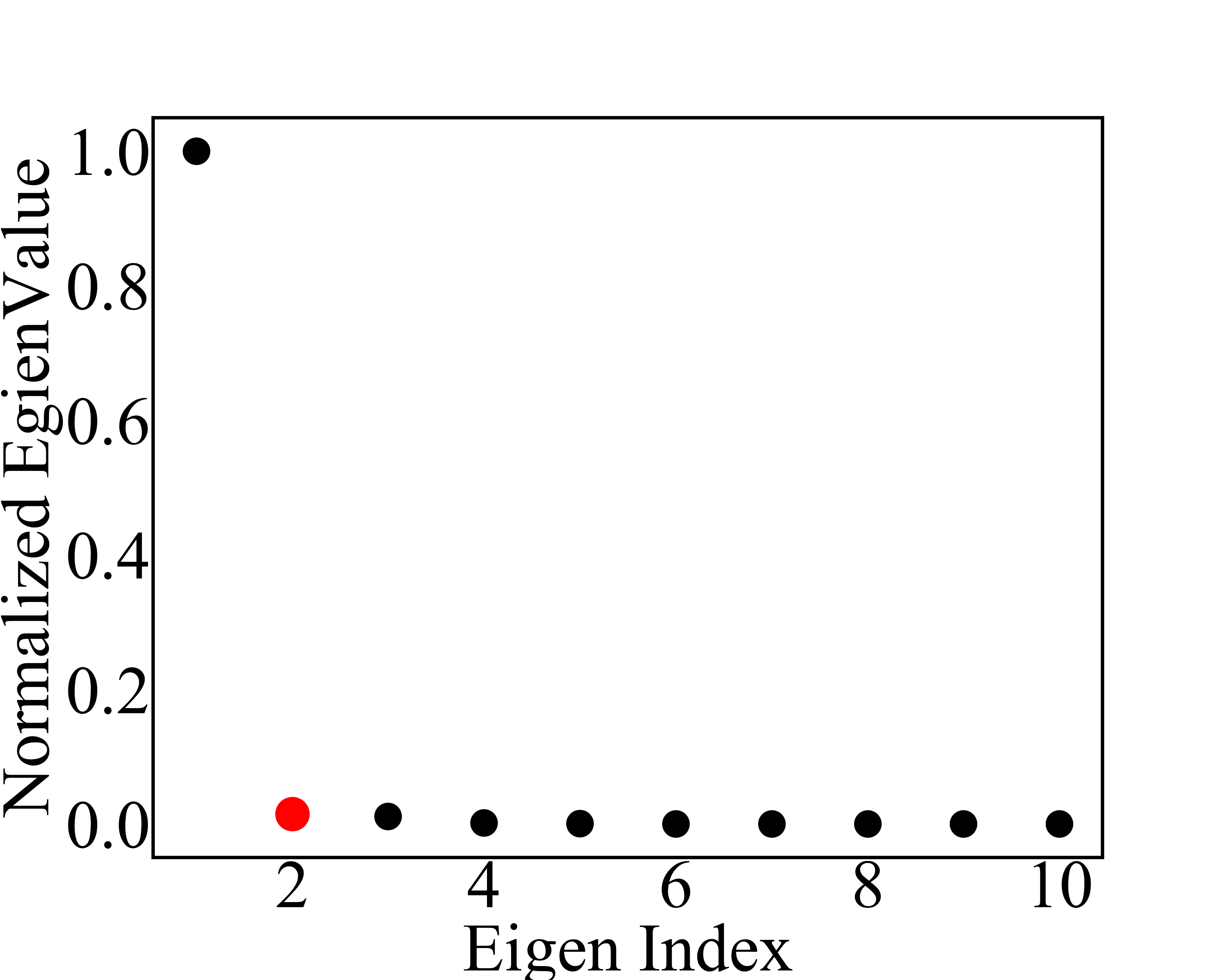}
        \caption{ \footnotesize{Elbow plot of \MM} }
        \label{fig:Elbow}
    \end{subfigure}
    \begin{subfigure}[b]{0.3\textwidth}
    \centering
    \captionsetup{justification=centering}
       \includegraphics[trim={0cm 3cm 3.5cm 10cm},width=\textwidth]{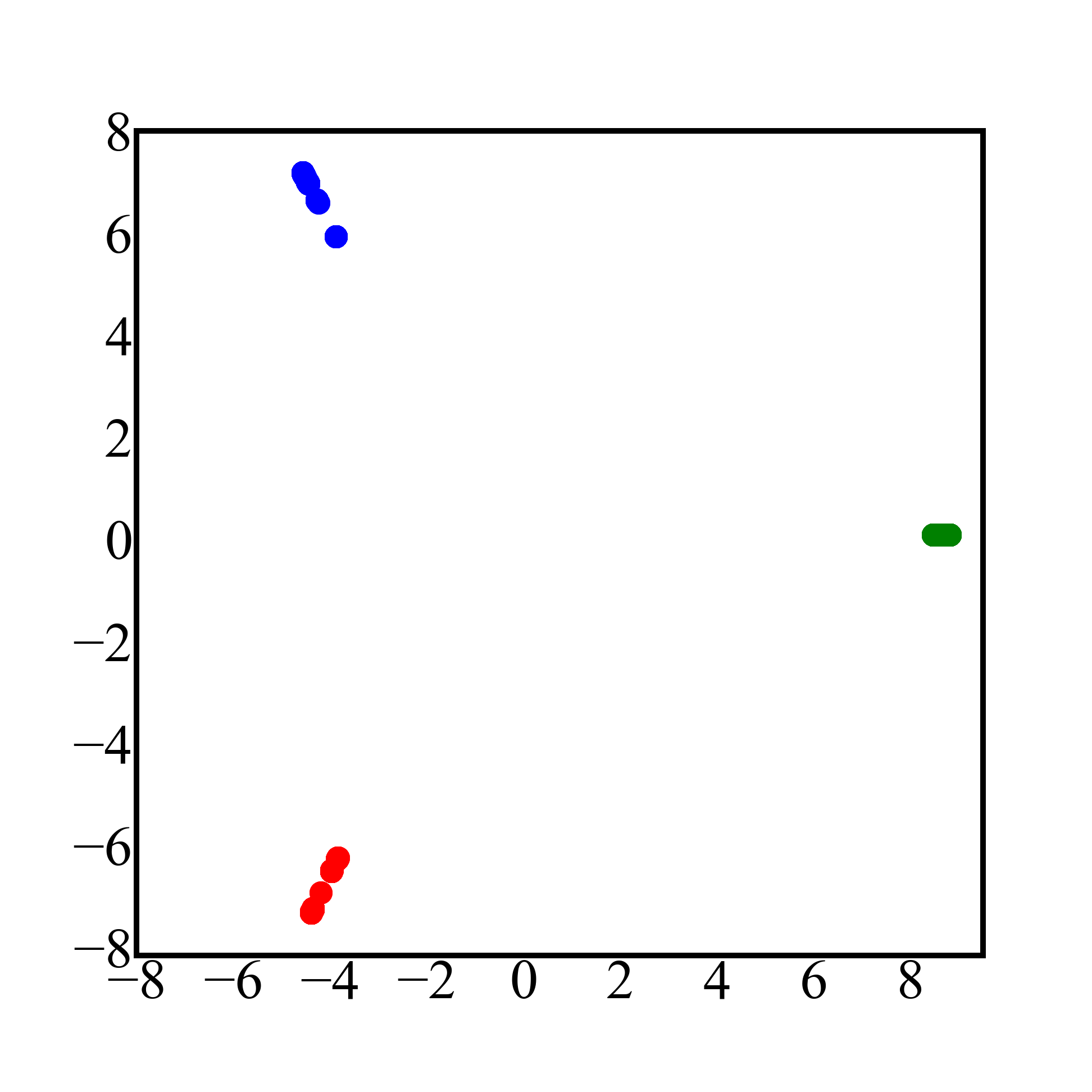}
        \caption{\footnotesize{MDS embedded \MM} }
        \label{fig:MM_embedding}
    \end{subfigure}
        \end{minipage}
    \captionsetup{singlelinecheck = false, justification=justified}
    \caption{ (a) Elbow plot of the \MM matrix when $d=1,\dots ,10$, for Set6, $N=512$ points. (b) The MDS embedding of \MM, for $3$ recovered clusters, when $d=2$. }\label{fig:embedding}
\end{figure}

We run the different methods for a varying number of clusters in the range of 2 to 7. Since we have access to ground truth, we know the optimal number of clusters is three, so investigating more than seven clusters is unnecessary. As mentioned, to find the optimal number of clusters, we compute the silhouette score and consider the number corresponding to peak of the silhouette score as the optimal number of clusters. Figure \ref{fig:optimalclustering} illustrates silhouette, rand index, mutual information, and v-measure scores of different methods when the number of clusters varies between 2 and 7 for Set6 with $N=512$ (for similar experiments on other sets, we refer to the supplementary material). Figure \ref{fig:Sallihoutte} shows that the silhouette score has its peak in 3 clusters in \DTMM, which is consistent with the peaks of different evaluation criteria in Figures \ref{fig:optimalclustering}\cref{fig:randindex,fig:mutualinfo,fig:vmeasure}.
However, the consistency in peak points between the different evaluation criteria is lost for the baseline methods.
Consequently, while the estimated optimal number of clusters obtained by \DTMM\ is almost always the same as the true optimal, this is not always the case for the baseline methods.

In the next evaluation, we assume all methods can compute the true optimal number of clusters, which is three. Table \ref{tab:result} shows the three performance criteria and the silhouette score for baseline and our method applied to all the datasets introduced in Table \ref{tab:sets}.
We observe that on all the datasets, our method outperforms the baseline methods. It also provides the highest silhouette score. 

\begin{figure}[t]
\begin{minipage}[b]{.9\linewidth}    \centering
\begin{subfigure}[b]{\textwidth}
    \captionsetup{justification=centering}\centering
        \includegraphics[trim={0cm 8cm 3cm 0cm},clip,width=0.8\textwidth]{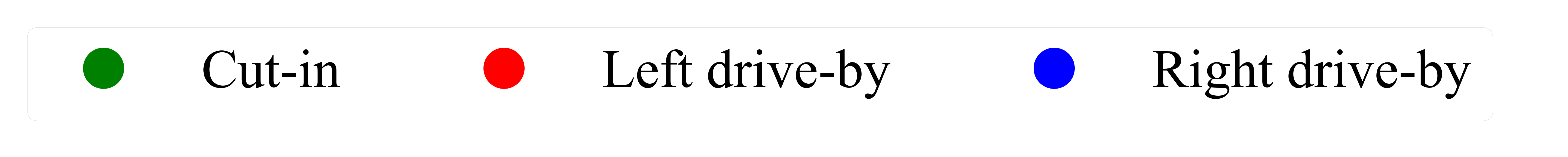}
    \end{subfigure}
    \begin{subfigure}[b]{0.3\textwidth}
    \captionsetup{justification=centering,skip=-0.2cm}
        \includegraphics[width=\textwidth]{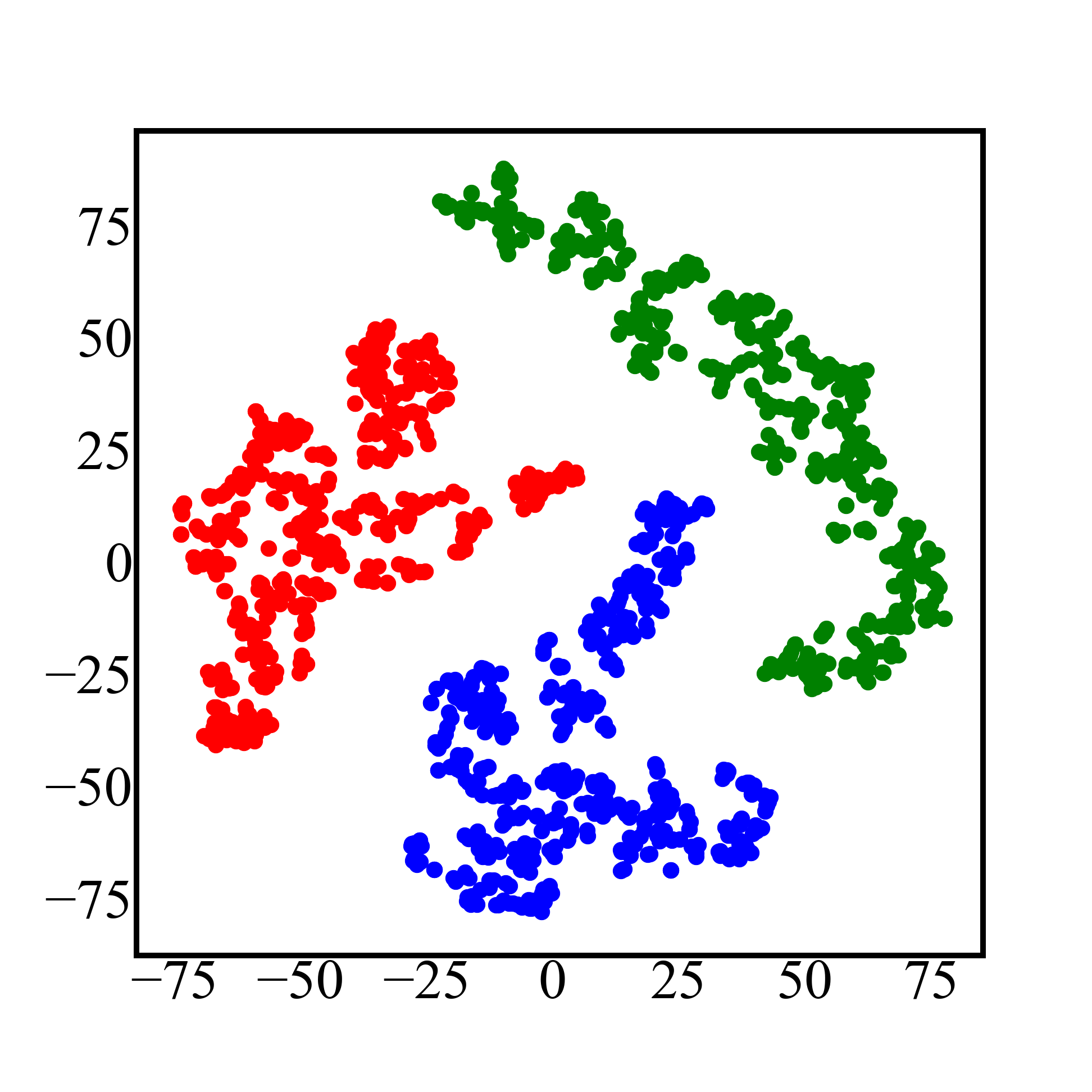}
        \caption{ \footnotesize{Ground truth} }
        \label{fig:datapoint}
    \end{subfigure}
    \begin{subfigure}[b]{0.3\textwidth}
    \captionsetup{justification=centering,skip=-0.2cm}
        \includegraphics[width=\textwidth]{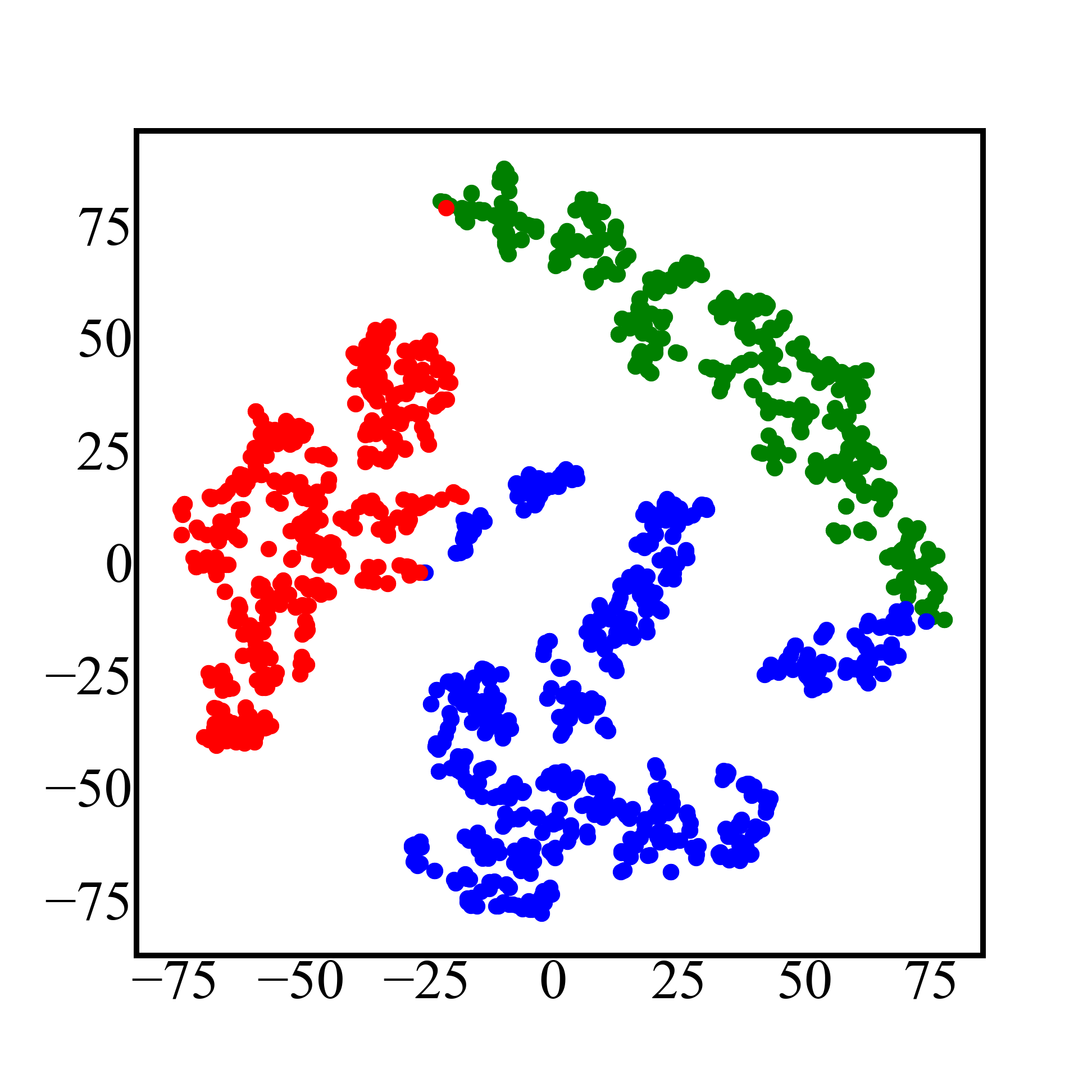}
        \caption{\footnotesize{Baseline1} }
        \label{fig:B1}
    \end{subfigure}
    \begin{subfigure}[b]{0.3\textwidth}
    \captionsetup{justification=centering,skip=-0.2cm}
        \includegraphics[width=\textwidth]{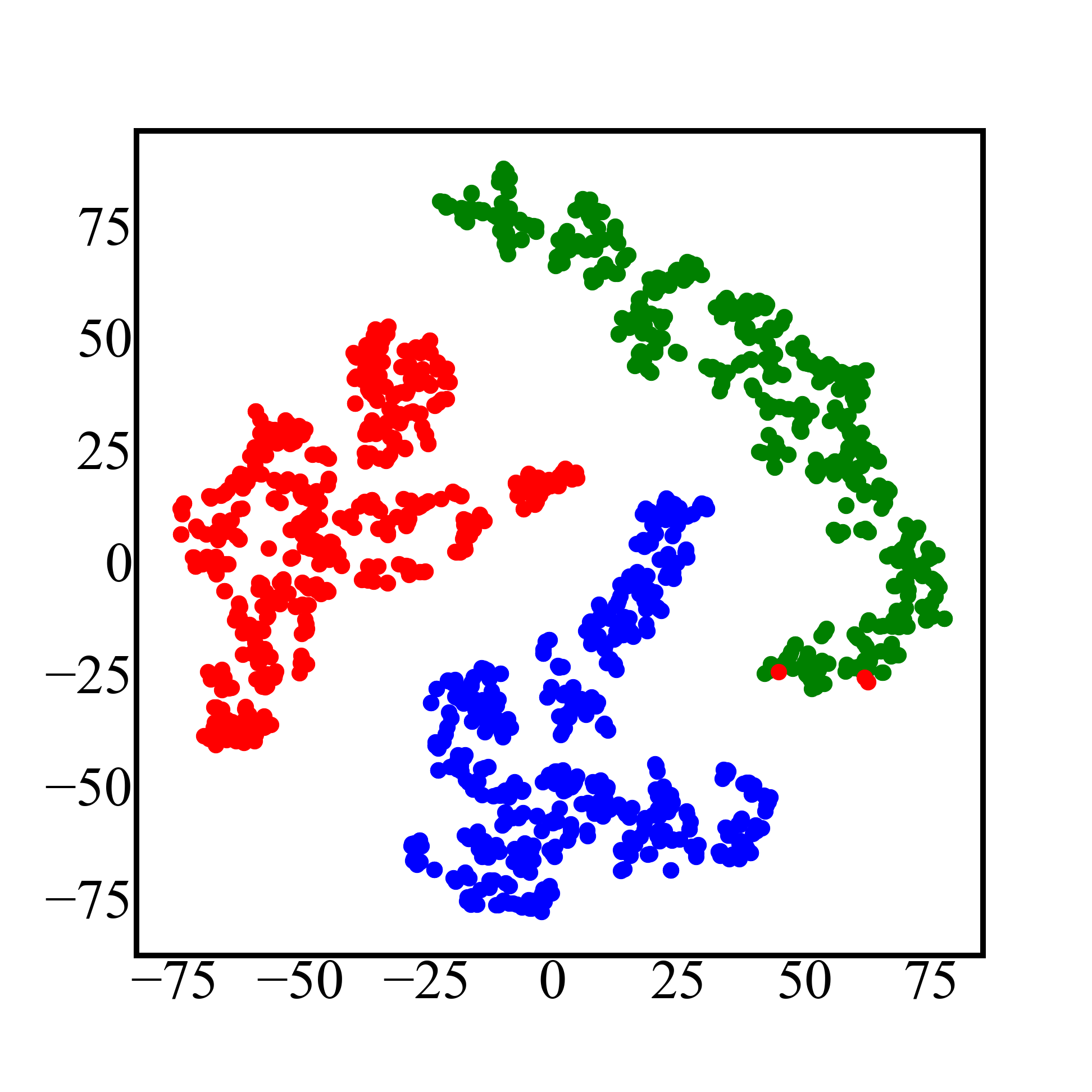}
        \caption{\footnotesize{Baseline2}  }
        \label{fig:B2}
    \end{subfigure}
    \begin{subfigure}[b]{0.3\textwidth}
    \captionsetup{justification=centering,skip=-0.2cm}
        \includegraphics[width=\textwidth]{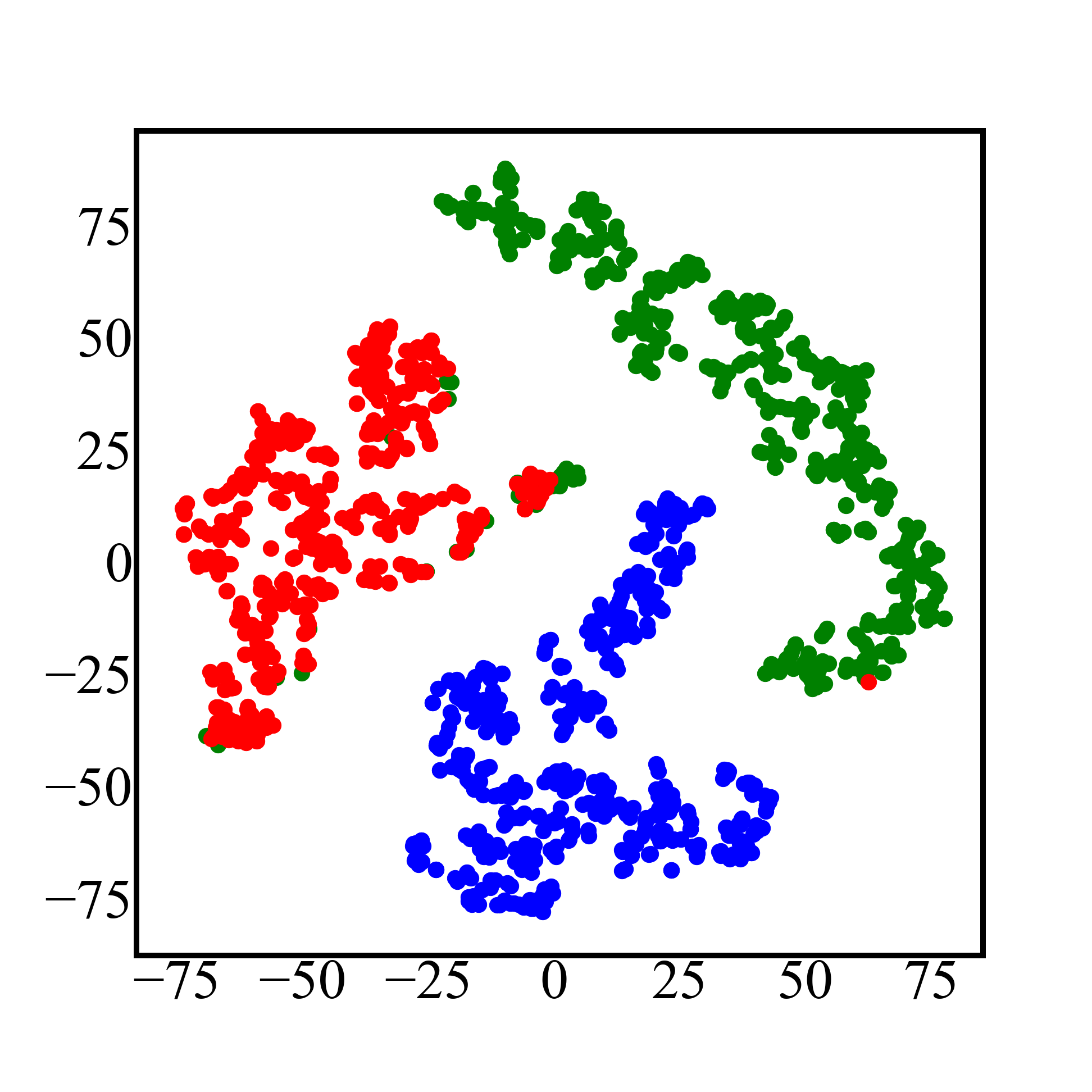}
        \caption{ \footnotesize{Baseline3}  }
        \label{fig:B3}
    \end{subfigure}
    \begin{subfigure}[b]{0.3\textwidth}
    \captionsetup{justification=centering,skip=-0.2cm}
        \includegraphics[width=\textwidth]{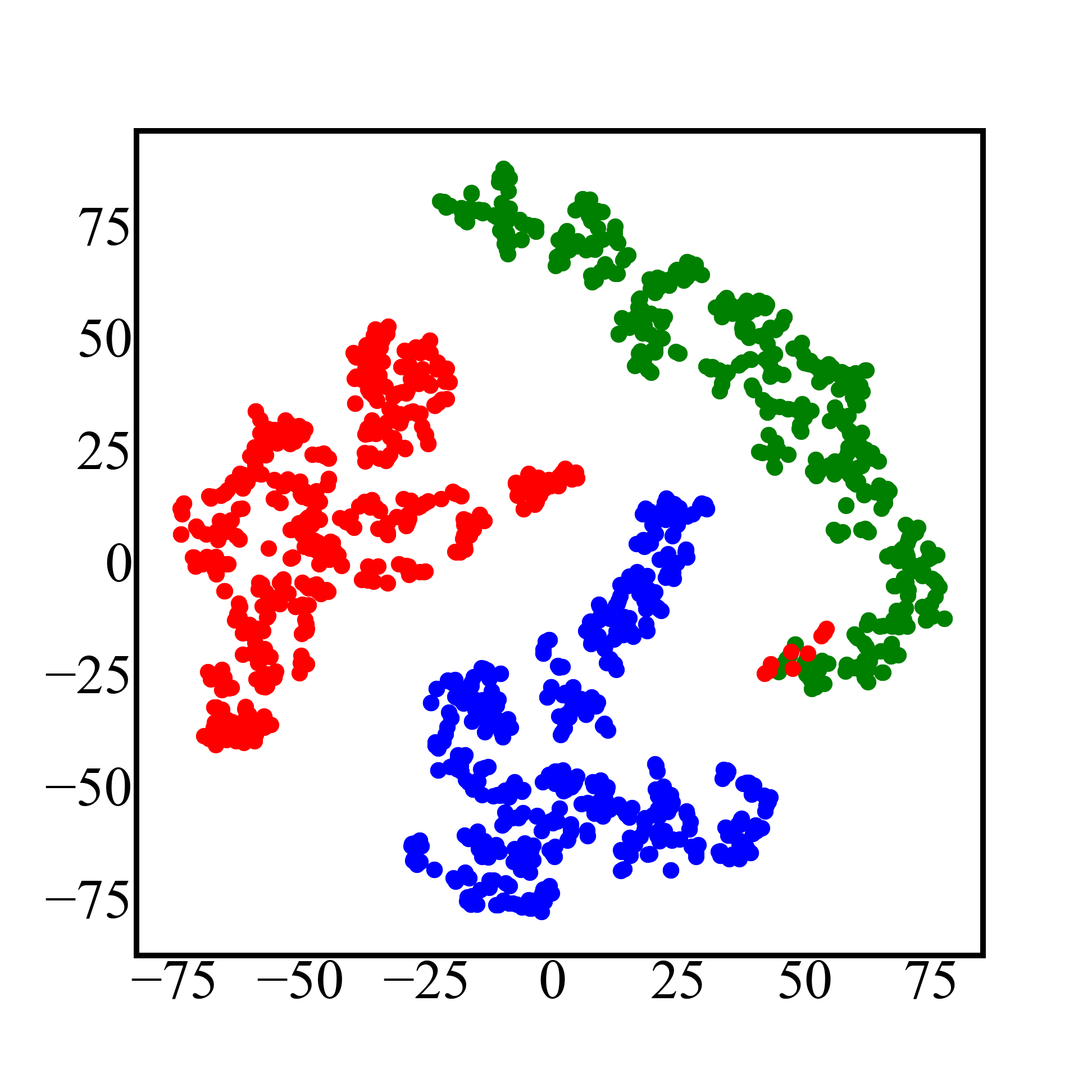}
        \caption{\footnotesize{Baseline4} }
        \label{fig:B4}
    \end{subfigure}
    \begin{subfigure}[b]{0.3\textwidth}
    \captionsetup{justification=centering,skip=-0.2cm}
        \includegraphics[width=\textwidth]{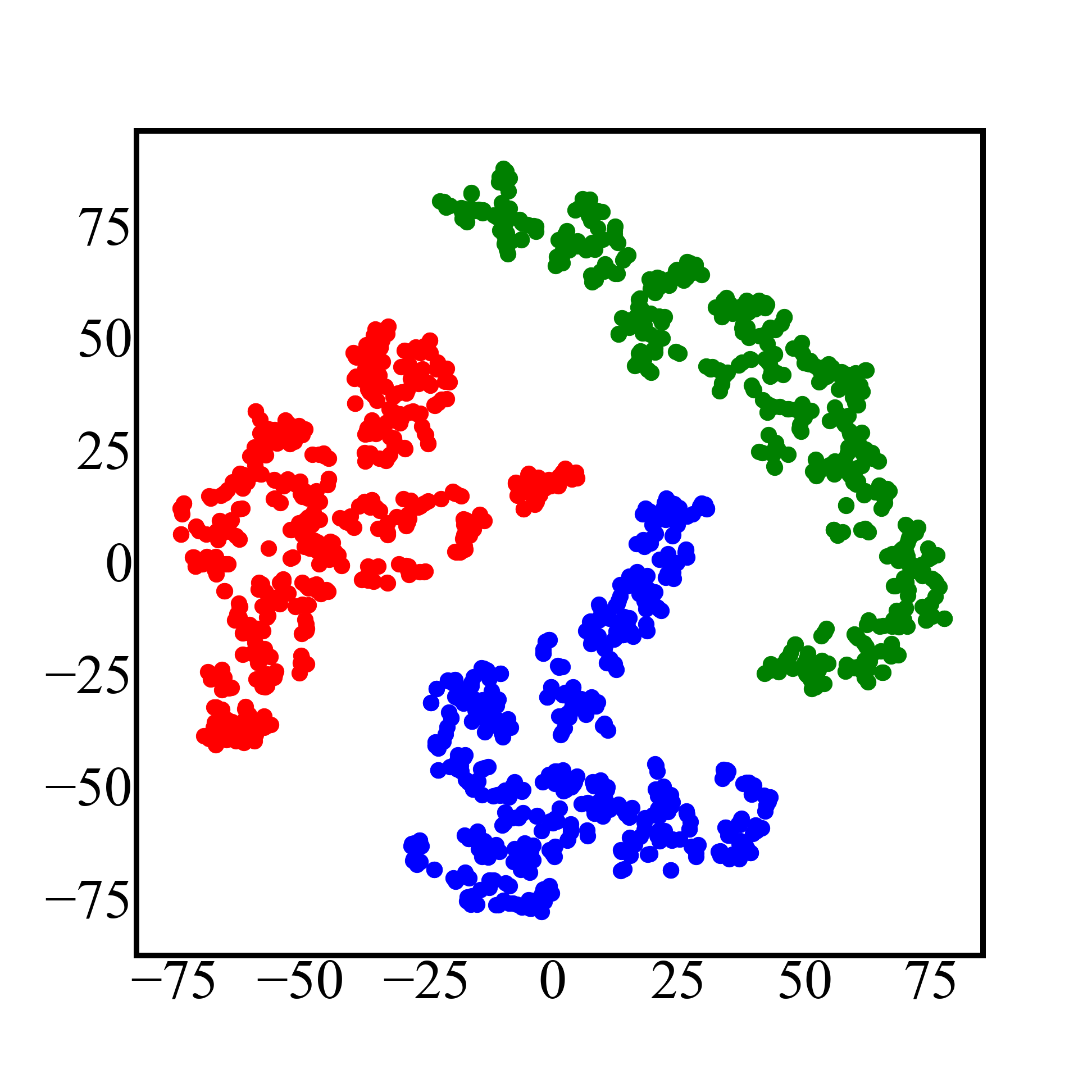} 
        \caption{ \footnotesize{\DTMM}  }
        \label{fig:DTMM}
    \end{subfigure}
        \end{minipage}
    \captionsetup{singlelinecheck = false, justification=justified}
    \caption{
    Illustration of clustering results of \DTMM\ and baseline methods for Set6 when $N=512$. For visualization purposes, the clusters are assigned to their closest ground truth clusters, and labeled accordingly.
    The embeddings are obtained by applying t-SNE after DTW.
    }
    \label{fig:result}
\end{figure}

Comparing the results of \DTMM\ with Baseline3 Baseline4, we can conclude that embedding the alignment-cost matrix using t-SNE proves superior to embedding with MDS, or not embedding at all.
This is probably due to that otherwise complex and shape-varying clusters now have clear boundaries between each other (Figure \ref{fig:result})
Then, comparing \DTMM\ with Baseline1, the use of \MM\ also proves advantageous, probably since it enabling us to extract the transitive relations and map the complex shape-varying clusters into very well separable dense clusters.
We conclude that the combination of t-SNE and \MM\ is the main contributing factor to the outstanding performance of our method by detecting the complex elongated patterns and preserving the locality.

Figure \ref{fig:result} illustrates the clustering results obtained by \DTMM\ along with the baselines, when applied to Set6, with $N=512$. For further illustrations on other sets, we refer to the supplementary material. Figure  \ref{fig:datapoint} shows the ground truth labels, whereas, in Figures \ref{fig:result}\cref{fig:B1,fig:B2,fig:B3,fig:B4,fig:DTMM},
the predicted labels by different methods are color-coded. Figure \ref{fig:B1} represents the Baseline1 clustering results. One can observe there are some points predicted as cut-in, which truly belong to two other clusters. Figure  \ref{fig:B2} shows the predicted label by Baseline2, in which we apply non-metric MDS. To determine the number of components needed, we utilize the cumulative explained variance ratio of $0.95$, which gives us to the number of components equal to three and better results than other choices. Figure \ref{fig:B3} shows the Baseline3 where we perform non-metric MDS with three components and embed the \MM\ matrix into a two-dimensional space. For Baseline4, we apply \MM\ right after DTW where  the result is shown in Figure \ref{fig:B4}. Finally, Figure \ref{fig:DTMM} illustrates the predicted labels by \DTMM, which is fully consistent with the ground truth labels, shown in Figure \ref{fig:datapoint}.

By choosing a low-dimensional space when embedding the \MM\ matrix, data becomes more separable, and we find a small $d$ by the Elbow trick. Experiments show that $\forall\ d \geq 1 $ \DTMM, results are stable.
Consequently, \DTMM\ does not require tuning any critical parameters, a very important aspect in unsupervised learning. 

Figure \ref{fig:embedding} shows the Elbow plot and the embedded data points when $d = 2$ for Set6, with N equal to 512.
Figure \ref{fig:Elbow} illustrates that the eigenvalues corresponding to the embedding drop very quickly at the second eigenvalue, shown by red mark. We consider this point as the proper value for $d$. Figure \ref{fig:MM_embedding} shows the embedded data points in the new space after clustering, whereas clusters labels are color-coded. The data points are very well separable.

\subsection{Analysis of GAN trajectories}

\begin{figure}[b]\centering
\begin{minipage}[b]{0.8\linewidth}  \centering
\begin{subfigure}[b]{\textwidth}
    \captionsetup{justification=centering} \centering
        \includegraphics[trim={0cm 0.9cm 0cm 0.5cm},clip,width=0.7\textwidth]{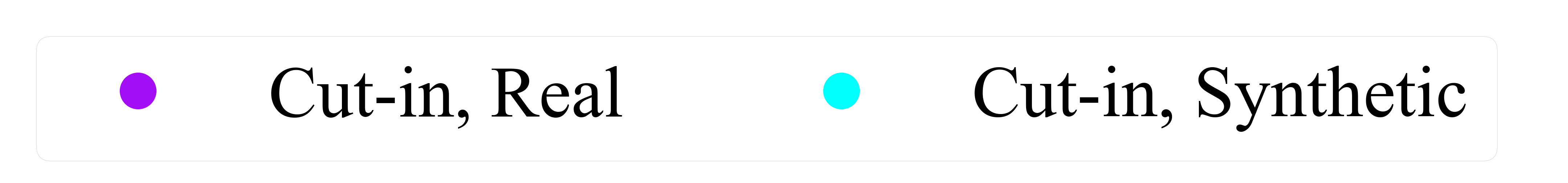}
    \end{subfigure}
    \begin{subfigure}[b]{0.3\textwidth}
    \captionsetup{justification=centering}
        \includegraphics[trim={4cm 4.5cm 3.2cm 4.5cm},width=\textwidth]{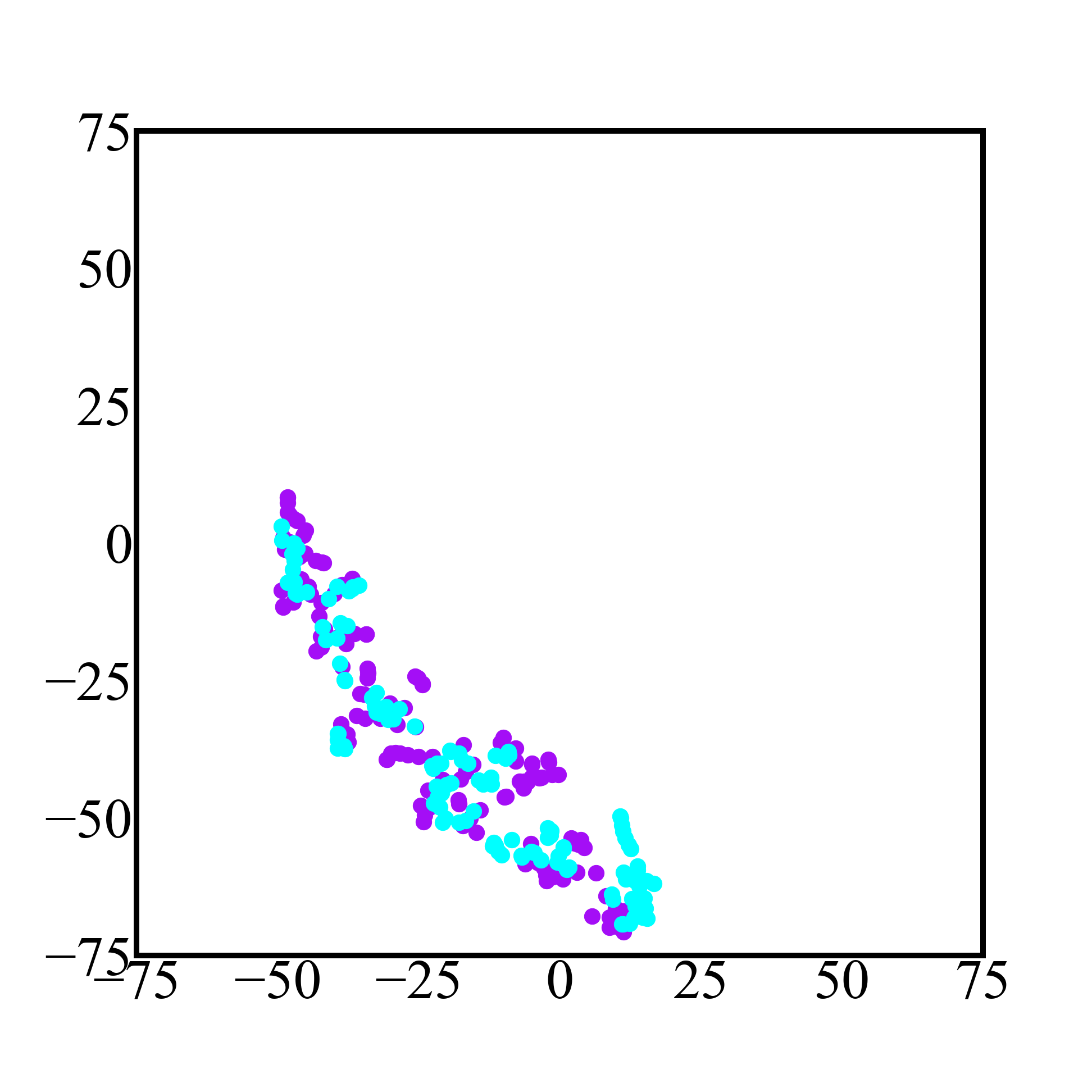}
        \caption{ \footnotesize{RecAE-GAN} }
        \label{fig:ae_gan_embeddings}
    \end{subfigure}
    \begin{subfigure}[b]{0.3\textwidth}
    \captionsetup{justification=centering}
       \includegraphics[trim={4cm 4.5cm 3.2cm 4.5cm},width=\textwidth]{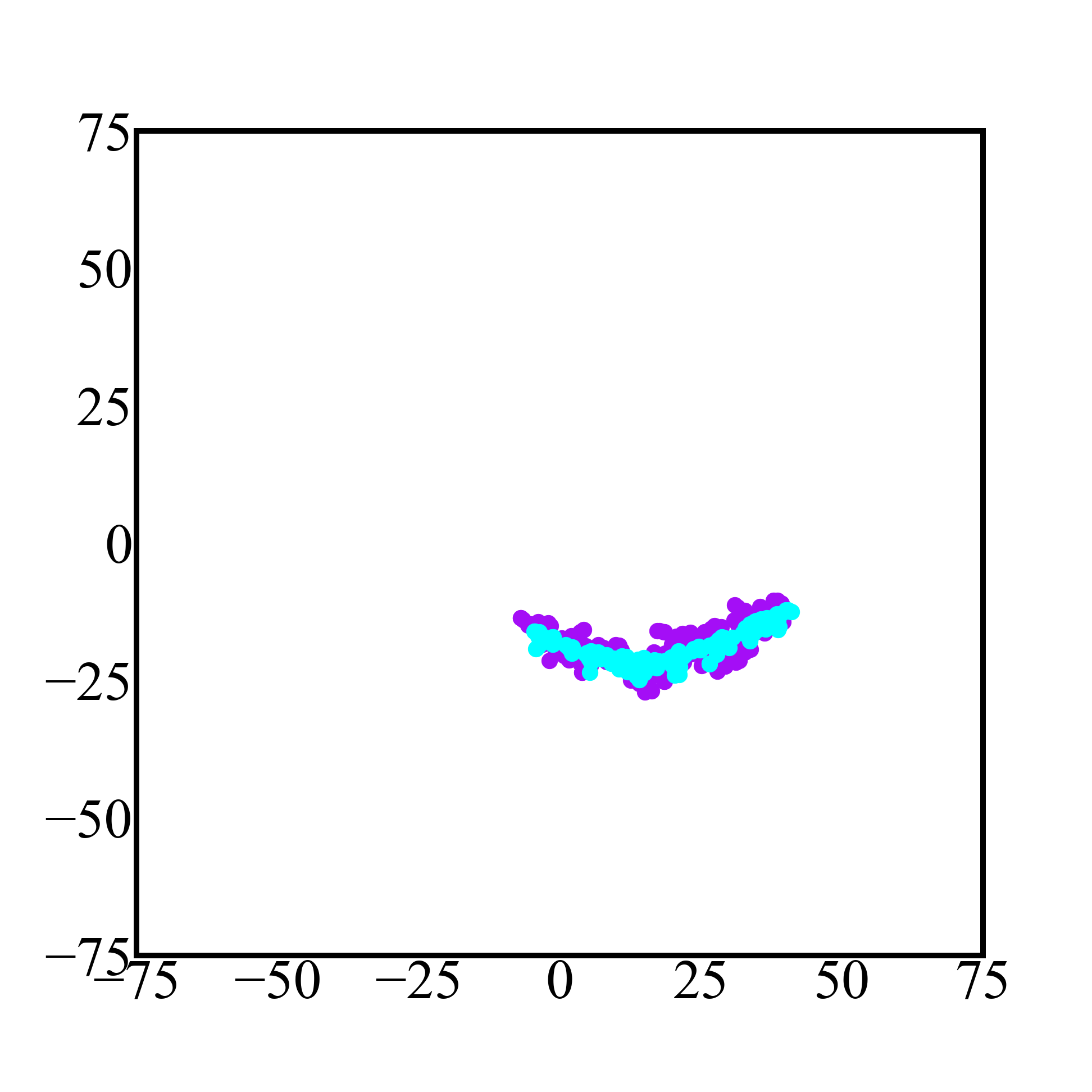}
        \caption{\footnotesize{WGAN} }
        \label{fig:wgan_embeddings}
    \end{subfigure}
        \end{minipage}
    \captionsetup{singlelinecheck = false, justification=justified}
    \caption{
        Embeddings for 128 real trajectories, as well as 128 additional synthetic trajectories, generated either by a RecAE-GAN or a RecAE-WGAN.
        The embeddings are obtained by applying t-SNE after DTW.
    }\label{fig:gan_embeddings}
\end{figure}

\begin{table}[]
\caption{
    Clustering results of \DTMM, when the amount of real data is small.
    Set5 and Set6 include extra synthetic trajectories, generated by a RecAE-GAN and a RecAE-WGAN, respectively.
}\label{tab:gan_results}
\centering
\resizebox{0.60\linewidth}{!}{
\begin{tabular}{|c|c|c|c|c|c|c|c|c|}
\hline
\multicolumn{3}{|c|}{Set4} & \multicolumn{3}{c|}{Set5} & \multicolumn{3}{c|}{Set6} \\ \hline
RI      & MI      & VM     & RI      & MI     & VM     & RI      & MI     & VM     \\ \hline
0.546   & 0.636   & 0.688  & 1.0     & 1.0    & 1.0    & 1.0     & 1.0    & 1.0    \\ \hline
\end{tabular}
}
\end{table}

We now proceed to evaluate the synthetically generated cut-in trajectories, described in Section \ref{sec:GAN_validation}. We investigate the performance of \DTMM\ when the amount of real data is limited, and then study how the synthetic data generated  by GANs can help in this case.
Specifically, we will focus on Set4, Set5, and Set6, wherein a small number of real-world trajectories for the cut-in driving scenario is available ($N = 256$).
Set5 and Set6 consist of the same real data as Set4, but are augmented with additional synthetic trajectories, generated by RecAE-GAN and RecAE-WGAN, respectively.

The synthetic trajectories are illustrated in Figure \ref{fig:gan_embeddings}, where t-SNE embeddings are plotted for real cut-in trajectories, as well as the added synthetic cut-in trajectories, generated by RecAE-GAN (a) and RecAE-WGAN (b).
From a visual inspection,
we can observe that both models generate synthetic trajectories that are consistent with the real data distribution.

We quantitatively demonstrate   how \DTMM\ can benefit from the generated trajectories, and at the same time, can validate their quality and consistency with the real trajectories.
Table \ref{tab:gan_results} shows the clustering results for Set4, Set5, and Set6.
Due to the limited amount of data for the cut-in scenarios in Set4, \DTMM\ fails to cluster the trajectories correctly in this set.
However, leveraging on the data augmentation provided by the RecAE-GAN or RecAE-WGAN, we can again achieve perfect performance for \DTMM, as shown by the results for Set5 and Set6.
To conclude, we notice that \DTMM\ may struggle when data is scarce, but the combination of \DTMM\ with Generative Adversarial Networks proves very effective in this setting.
Furthermore, the good performance on clustering Set5 and Set6 gives further evidence that the generated trajectories are of high quality.

\section{Conclusion}

We developed an unsupervised learning framework for the clustering of vehicle trajectories with varying lengths.
We validated the performance of the framework on real-world datasets collected from real driving scenarios and demonstrated its satisfactory performance. Our framework does not require fixing critical (hyper)parameters, a common issue with many other methods. 
In the following, we studied the critical case of lack of enough trajectories for one of the clusters, namely the cut-in scenarios. To address this issue, we employed Generative Adversarial Networks (GANs) (e.g., RecAE-GAN and RecAE-WGAN) in order to augment this minority cluster. Then, we employed the proposed clustering framework to serve as a validation tool to assess the quality of the generated trajectories, in particular, to investigate the consistency of the synthetic data and the real clusters.

\bibliographystyle{unsrt}  
\bibliography{references}  






\end{document}